%% file: acl_latex.tex
\pdfoutput=1
\documentclass[11pt]{article}

\usepackage[preprint]{acl}

\usepackage{times}
\usepackage{latexsym}
\usepackage{amsmath}
\usepackage{amssymb}
\usepackage{booktabs}
\usepackage{subcaption}
\usepackage{makecell}
\usepackage{float}

\usepackage[T1]{fontenc}

\usepackage[utf8]{inputenc}

\usepackage{microtype}

\usepackage{inconsolata}
\usepackage{multirow}

\usepackage{graphicx}

\newcommand{\baseRLVR}{Base-RLVR}
\newcommand{\passkRLVR}{Pass@$k$-RLVR}
\newcommand{\pkpo}{PKPO}

\newcommand{\jplagDiv}{JPlag-RLVR}
\newcommand{\vendiDiv}{Vendi-div}

\usepackage{xspace}
\usepackage{bm}

\usepackage{amsmath}
\usepackage{bm}
\usepackage{xspace}

\DeclareRobustCommand{\passat}[1]{%
  \ensuremath{\text{Pass@}#1}\xspace
}

\DeclareRobustCommand{\passk}{\passat{k}}

\newenvironment{itemize*}%
  {\begin{itemize}%
    \setlength{\itemsep}{0pt}%
    \setlength{\parsep}{0pt}%
    \setlength{\parskip}{0pt}%
    \setlength{\topsep}{0pt}%
  }%
  {\end{itemize}}

\title{Beyond \passk: Redundancy-Aware RLVR for Multi-Sample Code Generation}

\author{
  Florian Le Bronnec\normalfont\textsuperscript{1} \\
  \normalfont\texttt{florian.lebronnec@riken.jp}
  \And
  Alexandre Verine\normalfont\textsuperscript{2} \\
  \normalfont\texttt{alexandre.verine@ens.fr}
  \AND
  Rio Yokota\normalfont\textsuperscript{1} \\
  \normalfont\texttt{rio.yokota@riken.jp}
  \And
  Benjamin Negrevergne\normalfont\textsuperscript{3} \\
  \normalfont\texttt{benjamin.negrevergne@dauphine.psl.eu}
  \AND
  \normalfont\fontsize{11}{10}\selectfont
  \textsuperscript{1}RIKEN Center for Computational Science, Tokyo, Japan \\
  \normalfont\fontsize{11}{10}\selectfont
  \textsuperscript{2}École Normale Supérieure Paris, PSL University, Paris, France \\
  \normalfont\fontsize{11}{10}\selectfont
  \textsuperscript{3}LAMSADE, CNRS, Université Paris-Dauphine-PSL, Paris, France
}

\begin{document}
\maketitle

\begin{abstract}
    LLMs for code generation are commonly evaluated in repeated-sampling settings using \passk, where multiple candidate programs are executed against unit tests under a finite sampling budget. While recent verifier-based reinforcement learning (RLVR) methods improve executable correctness, how these objectives affect redundancy among sampled programs remains poorly understood. In this work, we study implementation-level redundancy in code generation using JPlag, a plagiarism-detection system for code. Across models and benchmarks, we show that correctness-only RLVR often concentrates generations around repeated implementations, whereas \passk-aware objectives maintain lower redundancy and improve larger-budget performance.

    Motivated by these observations, we augment RLVR with direct anti-redundancy rewards based on JPlag similarity. Across 3 models and 3 benchmarks, discouraging near-duplicate generations reliably improves finite-budget executable performance, often matching or outperforming specialized \passk-aware objectives.
\end{abstract}
\section{Introduction}

\paragraph{Code generation and \passk.}
When solving programming tasks with large language models (LLMs), it is common to generate multiple candidate programs until one successfully passes the task tests \citep{hui2024qwen25codertechnicalreport,zheng2025what,jiang2026survey}. This repeated-sampling setting is formalized through \passk, which is the probability that at least one of \(k\) sampled programs is correct \citep{chen2021evaluatinglargelanguagemodels,Guo_2025}.

\paragraph{RLVR for code generation.}

Because generated programs can be executed directly against tests, code generation is a natural setting for reinforcement learning with verifiable rewards (RLVR) \citep{li2022competition}. State-of-the-art coding models rely heavily on such post-training optimization to improve executable correctness \citep{grattafiori2024llama3herdmodels,olmo2026olmo3}. As coding benchmarks increasingly emphasize repeated-sampling evaluation, RL objectives have also shifted from optimizing isolated generations toward optimizing finite-budget executable success under \passk metrics \citep{tang2025optimizing,walder2025pkpo}.

\paragraph{Finite-budget \passk evaluation.}
Although these objectives improve executable performance under different sampling budgets, the sampled-program behaviors underlying these gains remain poorly understood. In particular, \passk measures whether at least one sampled program is correct, but does not distinguish between repeatedly generating near-duplicate implementations and covering a broader set of distinct implementation patterns. This distinction matters because the usefulness of repeated sampling changes with \(k\): larger sampling budgets can benefit not only from reliable generations, but also from reducing redundancy among generated programs. This motivates our first research question:

\medskip
\noindent
{\bfseries Q1.} \emph{When multi-sample performance improves, does the model use its sampling budget to produce less redundant solutions or to repeat near-duplicate implementations?}

\medskip

\paragraph{JPlag as a redundancy diagnostic.}
Executable correctness alone does not distinguish between repeatedly sampling near-duplicate programs and producing a broader set of implementations under a fixed sampling budget. To study redundancy among generated programs, we use JPlag, a program-similarity system originally developed for plagiarism detection \citep{prechelt2002jplag}. Unlike lexical overlap metrics, JPlag uses a programming-language-aware tokenization and matching procedure, making it substantially less sensitive to superficial edits such as variable renaming, formatting changes, or comments. This makes it a useful diagnostic for repeated implementations in multi-sample code generation.

Applying this diagnostic to RLVR methods, we find that objectives targeting larger-\(k\) \passk\ performance tend to produce sampled sets with lower redundancy and higher JPlag diversity.

\paragraph{Direct anti-redundancy training.}
These observations motivate our second research question:

\medskip
\noindent
{\bfseries Q2.} \emph{Can implementation redundancy itself serve as a useful optimization target for finite-budget code generation?}
\medskip

To investigate this, we train RLVR objectives with explicit group-level anti-redundancy rewards based on JPlag similarity. This simple intervention reliably improves finite-budget executable performance, across models and benchmarks, often matching or outperforming specialized \passk-aware objectives.
\paragraph{Contributions.}
Our contributions are as follows:
\begin{itemize*}
    \item We introduce JPlag as a practical diagnostic for implementation-level redundancy in multi-sample code generation.

    \item We show that \passk-aware RL objectives systematically maintain lower redundancy among sampled programs than correctness-only RLVR.

    \item We show that directly discouraging near-duplicate generations improves finite-budget executable performance and can match or outperform specialized \passk-aware objectives, without simply increasing sampling entropy or relying on superficial output variation.
\end{itemize*}

Overall, our results identify implementation redundancy as both a useful diagnostic and a practical optimization target for repeated-sampling code generation.

\section{Related Work}

\paragraph{Repeated-sampling evaluation for code generation.}
\passk measures the probability that at least one of \(k\) sampled programs solves a coding task \citep{chen2021evaluatinglargelanguagemodels,li2022competition}. This metric makes finite-budget sampling central to code-generation evaluation, but it reduces a sampled set to executable success. Two models can therefore obtain similar \passk while producing very different sampled sets: one may repeatedly generate near-duplicate solutions, while another may distribute samples across less redundant implementations. Our work studies this implementation-level redundancy directly.

\paragraph{\passk-aware RLVR.}
RLVR improves executable code generation by optimizing verifier feedback from unit tests or task checkers \citep{shao_deepseekmath_2024,li2022competition,gehring2024rlef}. Recent work adapts RLVR to repeated-sampling evaluation through objectives aligned with \passk \citep{tang2025optimizing,walder2025pkpo}. These methods are the closest methodological neighbors to our work: they change credit assignment over sampled groups to improve finite-budget executable success. We ask a complementary question: how do these objectives affect redundancy among the sampled programs, and can redundancy itself serve as a useful group-level training signal?

\paragraph{Diversity and redundancy in generated programs.}
Several studies show that post-training can improve generation quality while reducing diversity or concentrating the sampled distribution \citep{kirk2024understanding,le-bronnec-etal-2024-exploring}. Explicit diversity incentives have also been explored for reasoning tasks, particularly in mathematical settings \citep{hu2026diver}. For code generation, prior work has analyzed diversity using model-based or semantic evaluation pipelines \citep{lee-etal-2025-diversely}. Such approaches aim to capture broad behavioral or algorithmic differences between generated programs. We focus on a more specific notion: whether repeated-sampling RLVR produces near-duplicate implementations, and how this redundancy relates to finite-budget performance. While JPlag is not a full semantic-equivalence metric, it provides a practical code-aware signal that is robust to superficial edits and sufficiently informative to analyze and discourage repeated implementations.

\section{Background and Setup}
\label{sec:background}

\paragraph{Generation setup.}
For each programming problem \(x\), a model with parameters \(\theta\) defines a distribution over candidate programs \(y \sim \pi_\theta(\cdot \mid x)\). We study the multi-sample setting: for each prompt, we sample a set of \(n\) independent programs
\[
    Y^{(n)} = \{y_1,\ldots,y_n\}.
\]
Each program is executed against the task tests or checker, producing a binary correctness value \(c(x,y_i) \in \{0,1\}\).




\paragraph{\passk evaluation.}
We use \passk as the main executable evaluation metric throughout this work. For a task \(x\), \passk measures whether at least one of \(k\) sampled programs is correct:
\begin{equation}
    \label{eq:prompt-passk}
    \passk(x)\!
    =\!
    \mathbb{E}_{\!Y^{(k)} \sim \pi_\theta(\cdot \mid x)^k}\!
    \!\left[\!
        \max_{y_i \in Y^{(k)}} \! c(x,y_i)\!
        \right].
\end{equation}

In practice, we estimate this quantity from a larger set of \(n\) sampled programs. Let \(m = \sum_{y \in Y^{(n)}} c(x,y)\)
denote the number of correct samples. We then use the standard estimator
\begin{equation}
    \label{eq:passk-estimator}
    \widehat{\passk}(x)
    =
    1 -
    \binom{n-m}{k}
    \Big/
    \binom{n}{k},
\end{equation}
with value \(1\) when \(n-m < k\).

For a dataset \(\mathcal{X}=\{x_1,\ldots,x_N\}\), we report
\begin{equation}
    \label{eq:dataset-passk}
    \passk(\mathcal{X})
    =
    \frac{1}{N}
    \sum_{\ell=1}^{N}
    \widehat{\passk}(x_\ell).
\end{equation}

In the remainder of the paper, we simply write \passk or p@$k$ for the dataset-level metric when the evaluation set is clear from context.

\subsection{Verifier-Based RLVR for Code Generation}
\label{sec:rlvr-background}

Verifier-based reinforcement learning with verifiable rewards uses executable feedback to post-train code models. We write these methods in a group-sampling form: for each prompt \(x\), the policy samples a group \(Y^{(n)}=\{y_1,\ldots,y_n\}\), with each \(y_i \sim \pi_\theta(\cdot \mid x)\) drawn independently. A reward \(R(x,Y^{(n)})\) is then computed over the samples:

\begin{equation}
    \label{eq:general-rlvr-objective}
    \begin{aligned}
        J_{\mathrm{RLVR}}(\theta)
        =
        \mathbb{E}_{x}
        \mathbb{E}_{\substack{
            Y^{(n)} \sim \pi_\theta(\cdot \mid x)^n \!\!\!
        }}
        \left[
        R(x,Y^{(n)})
        \right].
    \end{aligned}
\end{equation}
In policy-gradient implementations, the group reward is transformed into per-sample advantages to assign credit within a sampled group and reduce gradient variance. Updates then take the form
\begin{equation}
    \label{eq:policy-gradient-advantage}
    \sum_{i=1}^{n}
    A(x,y_i,Y^{(n)})
    \nabla_\theta \log \pi_\theta(y_i \mid x),
\end{equation}
where the methods below differ mainly in how the advantage \(A(x,y_i,Y^{(n)})\) is constructed.

\paragraph{\baseRLVR.}
The standard correctness-only baseline uses executable correctness as a group reward by summing verifier outcomes:\begin{equation}
    \label{eq:correctness-reward}
    R_{\mathrm{corr}}(x,Y^{(n)})
    =
    \sum_{i=1}^{n}
    c(x,y_i).
\end{equation}
Under this objective, each correct generation contributes additively to the reward, so credit is assigned independently to successful samples within the group.

\paragraph{\passkRLVR.}
\citet{tang2025optimizing} instead consider the repeated-sampling evaluation setting directly, where success depends on whether at least one program in the sampled set is correct. For a sampled group \(Y^{(k)}\) of size \(k\),
\begin{equation}
    \label{eq:passk-rlvr-reward}
    R_{\mathrm{pass@}k}(x,Y^{(k)})
    =
    \max_{y_j \in Y^{(k)}} c(x,y_j).
\end{equation}
With \(r_j=c(x,y_j)\), the leave-one-out advantage is
\begin{equation}
    \label{eq:passk-rlvr-advantage}
    A_i
    =
    \max_j r_j
    -
    \max_{j \ne i} r_j .
\end{equation}

Thus a sample receives positive advantage only when removing it changes the group-level success outcome. We use the centered variant from \citet{tang2025optimizing} in all experiments.
\begin{table*}[t]
    \centering
    \scriptsize
    \input{tables/qualitative_jplag_examples}
    \caption{Qualitative examples illustrating how JPlag similarity compares with lexical overlap. All examples use the prompt: ``Write a function to find the maximum value in a given heterogeneous list.'' The generated programs differ in variable names, surface form, and control structure.\vspace{-0.2cm}}
    \label{tab:qualitative-jplag}
\end{table*}
\paragraph{\pkpo.}
\citet{walder2025pkpo} extend the same pass-at-\(k\) principle to groups with size \(n \geq k\). Rather than computing the advantage from a single sampled group, \pkpo{} averages the leave-one-out advantage of each sample over all \(k\)-subsets:
\begin{equation*}
    A_i^{\mathrm{PKPO}}
    =\frac{1}{\binom{n}{k}}
    \!
    \sum_{\substack{I \subseteq \{1,\ldots,n\}\\|I|=k,\ i\in I}}
    \!\!\!
    \Bigg(
    \max_{j\in I} r_j
    -\max_{j\in I\setminus\{i\}} r_j
    \Bigg).
\end{equation*}

This mirrors the standard \passk{} estimator eq.~\eqref{eq:passk-estimator} while reducing variance relative to relying on a single sampled group. In our experiments, we use their proposed \texttt{sloo\_minus\_one} estimator.

\subsection{Diversity and Redundancy Diagnostics}
\label{sec:jplag-diagnostics}

To study how RL objectives use a finite sampling budget, we measure redundancy among sampled programs using several complementary diagnostics. Existing approaches span a broad spectrum of tradeoffs. Lexical \(n\)-gram diversity measures local surface overlap between generations and is lightweight to compute, but remains sensitive to superficial edits. Embedding-based diagnostics such as the Vendi score \citep{friedman2023the} computed over code embeddings \citep{kryvosheieva2025efficientcodeembeddingscode} capture broader dispersion patterns, but are less directly interpretable in terms of repeated implementations.

\paragraph{JPlag diversity.}
Executable code generation provides a setting where redundancy can be analyzed more directly through program similarity. Our main diagnostic is therefore based on JPlag, a program-similarity system originally developed for plagiarism detection~\citep{prechelt2002jplag}. Compared to lexical overlap metrics, JPlag is substantially less sensitive to superficial edits such as variable renaming or formatting changes, making it better aligned with repeated implementations in generated code. At the same time, JPlag remains lightweight, interpretable, and scalable compared to semantic or execution-based similarity pipelines, making it practical for repeated-sampling RLVR analysis.

Table~\ref{tab:qualitative-jplag} illustrates this behavior on generated solution pairs for the same prompt. Programs with different variable names but similar control structure receive high JPlag similarity, while programs relying on different implementations receive lower similarity despite solving the same task.

For each prompt \(x\) and sampled set of programs \(Y^{(n)}\), we compute \(s(y_i,y_j) \in [0, 1]\) , the JPlag similarity between sampled programs \(y_i\) and \(y_j\). We define JPlag diversity (JDiv) as one minus the mean pairwise similarity:
\begin{equation}
    \label{eq:jplag-diversity}
    \text{JDiv}(Y^{(n)})
    \!=\!
    1\!-\!\frac{2}{n(n\!-\!1)}\!\!
    \sum_{1 \leq i < j \leq n}\!\!\!\!s(y_i,y_j).
\end{equation}

Larger values indicate lower redundancy among sampled programs. Appendix~\ref{app:jplag-cluster-analysis} additionally studies cluster-based diagnostics derived from JPlag similarity graphs. We also report lexical and embedding-based diversity diagnostics in our experiments. While correlated with JPlag diversity, these metrics capture partially different aspects of sampled-program variation (App.~\ref{app:metrics-discussions}, Fig.~\ref{fig:diversity-metric-correlations}).
\begin{figure*}[h]
    \centering
    \begin{subfigure}[t]{0.28\textwidth}
        \centering
        \includegraphics[width=\linewidth]{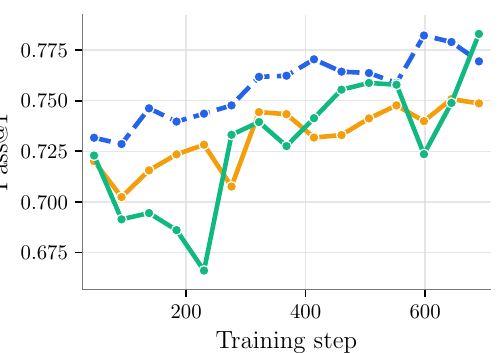}
        \caption{\passat{1}}
        \label{fig:baseline-training-dynamics-pass1}
    \end{subfigure}
    \hfill
    \begin{subfigure}[t]{0.28\textwidth}
        \centering
        \includegraphics[width=\linewidth]{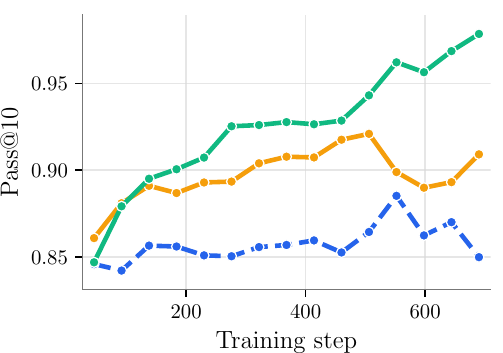}
        \caption{\passat{10}}
        \label{fig:baseline-training-dynamics-pass10}
    \end{subfigure}
    \hfill
    \begin{subfigure}[t]{0.385\textwidth}
        \centering
        \includegraphics[width=\linewidth]{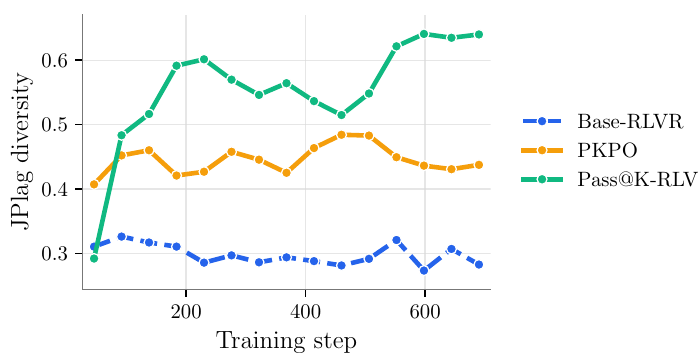}
        \caption{JPlag diversity}
        \label{fig:baseline-training-dynamics-diversity}
    \end{subfigure}
    \caption{Training dynamics for Qwen3-4B on MBPP. Correctness-only RLVR improves \passat{1} while reducing JPlag diversity, indicating increasing concentration on near-duplicate implementations. In contrast, pass@k-aware objectives systematically maintain higher diversity and primarily improve larger-budget performance (\passat{10}).}
    \label{fig:baseline-training-dynamics}
\end{figure*}

\section{RL Objectives Reshape Sample Redundancy}\label{sec:redundancy-analysis}

\paragraph{Same executable optimum, different finite-sample training signals.}
With binary verifier rewards, correctness-only RLVR and \passk-aware objectives optimize the same executable outcome and do not explicitly encourage implementation diversity. Their main difference lies in how training credit is assigned within sampled groups. Correctness-only RLVR reinforces each successful generation independently, whereas \passk-aware objectives reward generations according to whether they change the success of the sampled set. As a result, these objectives may induce qualitatively different generation behaviors during training. We study whether these differences manifest as systematic changes in redundancy among sampled programs.

\paragraph{Experimental setup.}
We compare how different RLVR objectives reshape executable performance and sampled-program redundancy after post-training from the same base models. We train Qwen3-4B, Qwen3-8B~\citep{yang2025qwen3technicalreport}, and Olmo3-7B~\citep{olmo2026olmo3} with correctness-only RLVR and \passk-aware objectives on MBPP~\citep{mbpp}, Code-Contest~\citep{li2022competition}, and TACO-Cobalt~\citep{taco-cobalt,taco-original}. All methods use the same training budget and sampling configuration. Evaluation uses \(200\) sampled generations per prompt from the test set.

\paragraph{Correctness-only RLVR concentrates repeated implementations.}
Across models and datasets, correctness-only RLVR tends to improve executable reliability while concentrating sampled generations around repeated implementations. Figure~\ref{fig:baseline-training-dynamics} illustrates this behavior for Qwen3-4B on MBPP: while \baseRLVR{} improves \passat{1}, JPlag diversity steadily decreases during training, indicating increasing concentration on near-duplicate solutions.

Table~\ref{tab:aggregate-baseline-diversity} shows that this trend extends beyond a single training run. The table aggregates prompt-level comparisons between each RL-trained model and its corresponding base model across all datasets and models, resulting in \(2745\) total prompt-level comparisons per method. Relative to the corresponding base model, \baseRLVR{} decreases JPlag diversity in \(57.2\%\) of comparisons and decreases Vendi score in \(65.3\%\) of comparisons. Although the mean JPlag-diversity change remains modest overall, the aggregate direction is consistent with correctness-only RLVR repeatedly reinforcing a narrower set of successful implementations during training.

\paragraph{\passk-aware objectives reduce redundancy among sampled programs.}
In contrast to correctness-only RLVR, \passk-aware objectives systematically maintain lower redundancy among sampled generations during training. In Table~\ref{tab:aggregate-baseline-diversity}, \passkRLVR{} increases JPlag diversity in \(66.9\%\) of comparisons and Vendi score in \(73.3\%\) of comparisons. \pkpo{} exhibits the same overall JPlag-diversity trend, increasing JPlag diversity in \(60.0\%\) of comparisons.

Together with the trajectories in Figure~\ref{fig:baseline-training-dynamics} and App.~\ref{app:rlvr-trajectories}, these results indicate that repeated-sampling objectives implicitly reduce redundancy among sampled programs, despite optimizing the same executable outcome as correctness-only RLVR. This effect emerges without any explicit diversity reward, suggesting that group-level \passk\ credit assignment changes how the finite sampling budget is used during training.

Taken together, these results suggest that repeated-sampling RL objectives differ not only in executable performance, but also in the redundancy structure of the sampled programs they produce. This raises a natural question: can directly discouraging redundant samples make finite sampling budgets more effective?
\begin{table}[h]
    \centering
    \resizebox{\linewidth}{!}{\input{tables/baselines_comparison/aggregated_results.tex}}

    \caption{Aggregate changes after RLVR post-training, comparing each post-trained model against its corresponding base model on the same prompt across all evaluated datasets and models. Each comparison corresponds to one prompt evaluated before and after post-training. For each metric, $\uparrow$\% and $\downarrow$\% report the fraction of comparisons where the post-trained model improves or decreases relative to the base model, while $\Delta$ reports the mean signed change. Correctness-only RLVR tends to improve executable performance while increasing redundancy among sampled programs, whereas repeated-sampling objectives more consistently improve both executable performance and JPlag diversity.}
    \label{tab:aggregate-baseline-diversity}
\end{table}

\section{Redundancy-Aware RLVR}
\label{sec:redundancy-aware-rlvr}
\begin{figure*}[t]
    \centering
    \begin{subfigure}[t]{0.325\textwidth}
        \centering
        \includegraphics[width=\linewidth]{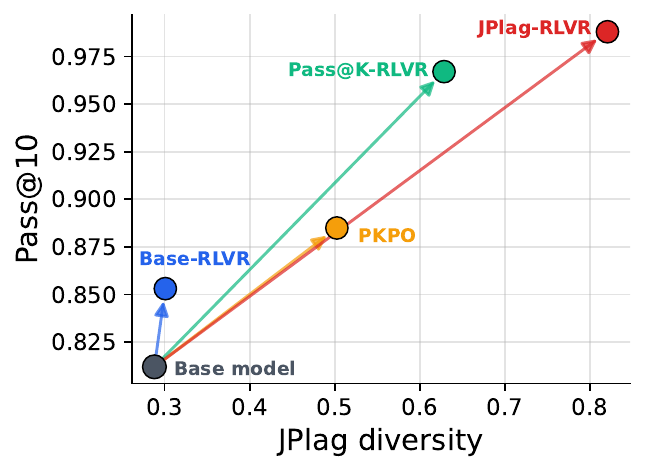}
        \caption{Qwen3-4B}
        \label{fig:tradeoff-mpbb-qwen3-4b}
    \end{subfigure}
    \hfill
    \begin{subfigure}[t]{0.325\textwidth}
        \centering
        \includegraphics[width=\linewidth]{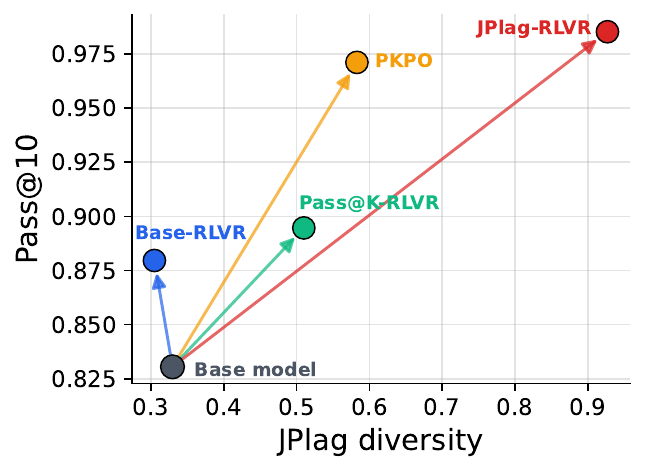}
        \caption{Qwen3-8B}
        \label{fig:tradeoff-mpbb-qwen3-8b}
    \end{subfigure}
    \hfill
    \begin{subfigure}[t]{0.325\textwidth}
        \centering
        \includegraphics[width=\linewidth]{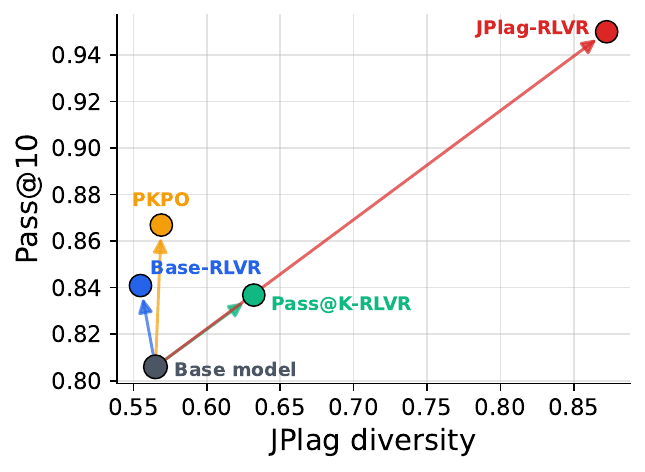}
        \caption{Olmo3-7B}
        \label{fig:tradeoff-mbpp-olmo}
    \end{subfigure}
    \caption{Tradeoff between repeated-sampling performance and redundancy on MBPP. The y-axis reports \passat{10}; the x-axis reports JPlag diversity, where higher values indicate less plagiarism-style redundancy. Arrows show movement from the base model to each training objective. Methods that move up and right improve multi-sample success while reducing redundancy.}
    \label{fig:tradeoff}
\end{figure*}

\subsection{Proposed Method}

We answer this question with a minimal intervention: augment the standard executable RLVR objective with a group-level anti-redundancy reward. The goal is not to reward diversity in isolation, but to discourage sampled groups from containing near-duplicate implementations while preserving executable correctness as the primary training signal.
\paragraph{Direct anti-redundancy training.} For a sampled group \(Y^{(n)}=\{y_1,\ldots,y_n\}\), we define a scalar anti-redundancy reward \(R_{\mathrm{div}}(x,Y^{(n)})\) from pairwise relations between generated programs. We combine it with the correctness reward \(R_{\mathrm{corr}}\) from Equation~\ref{eq:correctness-reward}:
\begin{equation}
    \label{eq:diversity-rlvr-reward}
    \begin{aligned}
        R(x,Y^{(n)})
         & =
        R_{\mathrm{corr}}(x,Y^{(n)}) \\
         & \quad+
        \lambda_{\mathrm{div}}
        R_{\mathrm{div}}(x,Y^{(n)}),
    \end{aligned}
\end{equation}

where \(\lambda_{\mathrm{div}}\) controls the strength of the anti-redundancy term.

\paragraph{Marginal contribution within sampled groups.} Because \(R_{\mathrm{div}}\) is group-level, its training signal should depend on each sample's marginal contribution to the sampled set. We therefore use the same leave-one-out construction as the repeated-sampling objectives in Section~\ref{sec:rlvr-background}. For the diversity component, the advantage of sample \(i\) is
\begin{equation}
    \label{eq:diversity-advantage}
    A_i^{\mathrm{div}}
    =
    R_{\mathrm{div}}(x,Y^{(n)})
    -
    R_{\mathrm{div}}(x,Y^{(n)}_{-i}),
\end{equation}
where \(Y^{(n)}_{-i}=Y^{(n)}\setminus\{y_i\}\). This assigns positive advantage to samples that make the group less redundant and negative advantage to samples that mainly duplicate other generations.

\paragraph{JPlag-RLVR.}
Our main instantiation uses JPlag-diversity from Equation~\ref{eq:jplag-diversity} as \(R_{\mathrm{div}}\). This yields a simple redundancy-aware RLVR objective: retain the executable reward, but add an explicit penalty against repeated samples within $Y^{(n)}$. We also evaluate simpler lexical and embedding-based variants as controls in Section~\ref{sec:ablations}.

\subsection{Main Results}

\begin{table*}[t]
    \centering
    \input{tables/main_results}

    \caption{Main executable performance and redundancy diagnostics across models, datasets, and training objectives. \passat{1}, \passat{10}, and \passat{100} report finite-budget executable success. JDiv. denotes JPlag diversity, 1gDiv. denotes 1-gram diversity, and Vendi denotes embedding-space diversity; higher is better for all three redundancy diagnostics. A star (*) marks methods significantly better than all non-diversity baselines under a paired bootstrap test over prompts (\(p < 0.05\)).\vspace{-0.35cm}}
    \label{tab:main-results}
\end{table*}

\paragraph{Experimental setup.}
We use the experimental setup as in Section~\ref{sec:redundancy-analysis}. Additional training details, including optimization budgets, hyperparameters, and dataset splits, are provided in Appendix~\ref{sec:implementation-details}.

\paragraph{Direct anti-redundancy improves finite-budget performance.}
Table~\ref{tab:main-results} shows that directly optimizing anti-redundancy is effective despite its simplicity. Across models and benchmarks, \jplagDiv{} often matches or outperforms specialized \passk-aware objectives while producing substantially lower redundancy among sampled programs. The gains are especially strong on MBPP, where \jplagDiv{} achieves the best \passat{1} and \passat{10} results for all three models, and on TACO, where it gives the strongest results for Qwen3-4B and Qwen3-8B.

For example, on MBPP with Qwen3-8B, \jplagDiv{} improves \passat{1} from \(79.1\) to \(90.0\) and \passat{10} from \(88.0\) to \(98.5\), while increasing JPlag diversity from \(0.305\) to \(0.927\). On TACO with Qwen3-8B, it improves \passat{10} from \(48.2\) to \(55.5\) and \passat{100} from \(58.5\) to \(68.7\), again with the highest JPlag diversity among all methods.

\paragraph{Anti-redundancy reshapes the performance--redundancy tradeoff.}
Figure~\ref{fig:tradeoff} summarizes this effect on MBPP. While the primary objective remains improved repeated-sampling performance, the JPlag-diversity axis provides an interpretable view of how different RL objectives use the sampling budget. Relative to correctness-only RLVR, \jplagDiv{} achieves higher executable performance together with lower measured redundancy among sampled programs. This supports the central hypothesis of the paper: directly discouraging near-duplicate generations can make finite sampling budgets more effective.

\paragraph{Gains are broadly distributed across prompts.}
The improvements from \jplagDiv{} are not driven by a few outlier tasks. Table~\ref{tab:aggregate-baseline-diversity} shows that \jplagDiv{} improves the largest fraction of prompt-level comparisons for \passat{1}, \passat{10}, and \passat{100}, while also giving the strongest average executable gains. These gains coincide with the largest increase in JPlag diversity. Appendix~\ref{app:jplag-cluster-analysis} further shows that \jplagDiv{} generally increases correct-only JPlag diversity and cluster dispersion, suggesting that anti-redundancy training increases useful variation rather than merely producing diverse incorrect outputs.

\subsection{Ablations and Analysis}
\label{sec:ablations}
\begin{figure*}[t]
    \centering
    \begin{subfigure}[t]{0.28\textwidth}
        \centering
        \includegraphics[width=\linewidth]{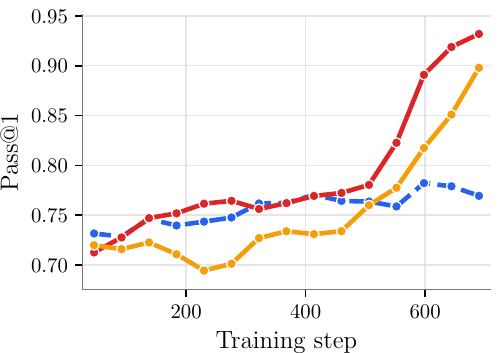}
        \caption{\passat{1}}
        \label{fig:diversity-coef-history-pass1}
    \end{subfigure}
    \hfill
    \begin{subfigure}[t]{0.28\textwidth}
        \centering
        \includegraphics[width=\linewidth]{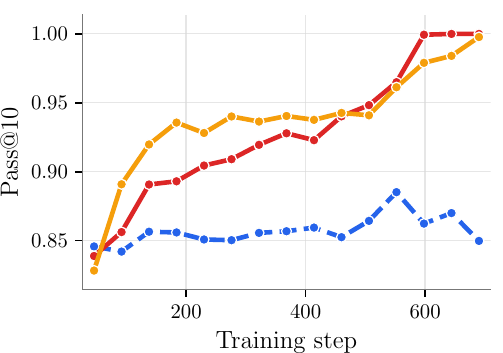}
        \caption{\passat{10}}
        \label{fig:diversity-coef-history-pass10}
    \end{subfigure}
    \hfill
    \begin{subfigure}[t]{0.385\textwidth}
        \centering
        \includegraphics[width=\linewidth]{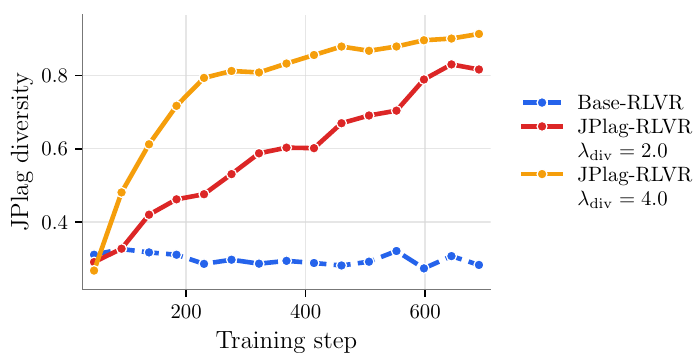}
        \caption{JPlag diversity}
        \label{fig:diversity-coef-history-diversity}
    \end{subfigure}
    \caption{Effect of the \jplagDiv{} coefficient during training for Qwen3-4B on MBPP. Larger diversity weight produces a stronger increase in JPlag diversity and, in this run, also improves \passat{10}.\vspace{-0.1cm}}
    \label{fig:diversity-coef-history}
\end{figure*}

\paragraph{Simpler anti-redundancy rewards partially help.}
We compare JPlag-based rewards with simpler group-level anti-redundancy signals in shorter experiments on Qwen3-4B on MBPP (Appendix~\ref{app:ablations-other-rewards}). We evaluate entropy regularization, lexical 1-gram diversity, and Vendi-embedding-based diversity rewards. Among these alternatives, 1-gram diversity gives the strongest repeated-sampling gains, suggesting that even simple lexical anti-redundancy signals can improve finite-budget performance. However, \jplagDiv{} gives the strongest redundancy diagnostics and the best \passat{100}, indicating that code-structure-aware similarity provides a more effective signal for reducing repeated implementations. Entropy and Vendi-based rewards are weaker in this focused comparison.

\paragraph{Anti-redundancy is not just higher sampling entropy.}
Increasing sampling temperature provides a simple test of whether the gains come merely from more random generations at inference time. As temperature increases from \(T=1\) to \(T=2\), \baseRLVR{} achieves higher \passat{100} and lower measured redundancy, but these improvements remain substantially smaller than those obtained with explicit anti-redundancy training and come at the cost of lower \passat{1}. In particular, \jplagDiv{} achieves much larger gains in \passat{10} and \passat{100} while further reducing redundancy. This suggests that effective anti-redundancy training is not equivalent to simply increasing sampling entropy.

\paragraph{Anti-redundancy gains are not explained by trivial verbosity.}
A possible concern is that anti-redundancy rewards could be satisfied through superficial reward hacking, such as longer completions, markdown artifacts, or non-executable text rather than genuinely different programs. We do not find evidence for this explanation: both raw completion lengths and extracted executable-code lengths remain broadly comparable across methods (Appendix~\ref{app:completion-lengths}). This suggests that the gains are not simply caused by producing longer or superficially modified outputs.

\paragraph{Redundancy reduction must be paired with correctness.}
Optimizing anti-redundancy without executable correctness is also insufficient. As a sanity check, we train Qwen3-4B on MBPP using only the JPlag diversity reward, removing the correctness term. Appendix~\ref{app:diversity-only} shows that this objective rapidly increases JPlag diversity, but executable performance collapses. Thus, lower redundancy is useful only when coupled with the verifier reward: the goal is not diversity in isolation, but non-redundant correct generations under a finite sampling budget.

\paragraph{Balancing correctness and redundancy.}
Figure~\ref{fig:diversity-coef-history} studies the effect of the anti-redundancy coefficient on Qwen3-4B for MBPP. Compared with \baseRLVR{}, both redundancy-aware runs increase JPlag diversity throughout training and improve \passat{10}. Larger coefficients reduce redundancy more strongly, but can slightly weaken executable reliability despite higher diversity. This illustrates the central tradeoff of redundancy-aware RLVR: anti-redundancy is useful as a controlled training signal, not as an objective to maximize independently of correctness.

\paragraph{Redundancy-aware training remains effective across random seeds.}
The improvements from \jplagDiv{} persist across independent training runs. We evaluate five seeds for Qwen3-4B on MBPP and TACO, covering both a relatively high-success short-form benchmark and a more difficult competitive-programming benchmark. Across these runs, \jplagDiv{} maintains substantially higher JPlag diversity than the baselines and achieves the strongest average repeated-sampling executable performance. Its variance is broadly comparable to the other RLVR objectives, suggesting that the anti-redundancy reward improves finite-budget behavior without introducing additional instability in these experiments.

\begin{table}[t]
    \centering
    \scriptsize
    \resizebox{\columnwidth}{!}{\input{tables/temperature_mpbb}}
    \caption{Temperature ablation for Qwen3-4B on MBPP. Increasing the sampling temperature for \baseRLVR{} trades lower \passat{1} for higher \passat{10}, higher \passat{100}, and higher JPlag diversity. Explicit anti-redundancy rewards still achieve substantially stronger multi-sample performance.\vspace{-0.1cm}}
    \label{tab:temperature-mpbb}
\end{table}

\section{Conclusion}

We studied redundancy as a hidden axis of repeated-sampling RLVR for code generation. Our results show that objectives with similar executable rewards can induce different sampled-program behavior: correctness-only RLVR often concentrates generations around repeated implementations, whereas \passk-aware objectives tend to maintain lower redundancy and improve larger-budget performance.

We then showed that this redundancy signal can be used directly. A simple JPlag-based anti-redundancy objective generally improves finite-budget \passk performance, often matching or outperforming specialized \passk-aware methods while increasing useful variation among successful generations.

Overall, these results suggest that finite-budget code generation should not be understood only through marginal correctness. How a model allocates its samples across repeated or distinct implementations is itself a useful diagnostic and a practical optimization target.

\clearpage

\section*{Limitations}

\paragraph{JPlag as a redundancy diagnostic.}
Our analysis relies on JPlag as a practical diagnostic for implementation-level redundancy in generated code. Compared with lexical overlap metrics, JPlag is substantially less sensitive to superficial edits such as variable renaming or formatting changes, making it useful for studying repeated implementations in sampled code generations. However, it remains an approximate structural similarity measure rather than a semantic equivalence oracle: two programs can implement the same underlying algorithm while receiving low JPlag similarity, and conversely structurally similar programs may differ semantically. Nevertheless, our results show that even this relatively simple structural signal is already useful both as a diagnostic and as a training objective, suggesting that richer program-analysis or semantic-equivalence tools could further improve redundancy-aware training in future work.

\paragraph{Scale and benchmark scope.}
Our experiments are limited to three open-weight models and competitive-programming-style benchmarks involving relatively short executable programs. Because RLVR training with repeated sampling, sandbox execution, and pairwise similarity computation is computationally expensive, we do not extensively explore larger-scale models, broader hyperparameter sweeps, or seed variation across all model--dataset settings. We include a targeted multi-seed evaluation for Qwen3-4B on MBPP and TACO, but broader stability studies remain future work. It also remains unclear whether the same redundancy patterns hold for long-horizon software engineering tasks, multi-file repositories, or agentic coding settings.

\paragraph{Dependence on executable verifiers.}
Like other RLVR methods, our approach depends on executable reward signals derived from tests or task checkers. Weak or incomplete test suites can reward programs that satisfy the verifier without fully solving the intended task. More generally, our framework applies most naturally to domains where correctness can be checked automatically.

\paragraph{Potential risks.}
By improving finite-budget code generation, our method could also make repeated-sampling attacks more effective by reducing the number of attempts needed to elicit harmful or insecure code from models, although we do not investigate this behavior. Like other verifier-based methods, it may also exploit incomplete tests rather than satisfy the intended specification. We restrict our experiments to standard programming benchmarks and sandboxed execution.

\section*{Acknowledgements}

This work was supported by the French government under the management of Agence Nationale de la Recherche as part of the ``Investissements d'avenir'' program, reference ANR-19-P3IA-0001. This work was granted access to the HPC resources of IDRIS under the allocations 2025-AD011014053R2, 2025-A0181016159, and 2025-AD011014022R3 made by GENCI.

\bibliography{references}

\clearpage
\onecolumn
\appendix
\renewcommand{\thefigure}{\thesection.\arabic{figure}}
\renewcommand{\thetable}{\thesection.\arabic{table}}

\section{Random Seed Variation}
\label{app:seed-variation}

To evaluate robustness across training runs, we additionally perform five independent runs for Qwen3-4B on both MBPP and TACO. These benchmarks provide complementary regimes, ranging from relatively high-success short-form problems (MBPP) to more difficult competitive-programming tasks (TACO).

Table~\ref{tab:seed-variation} shows that the qualitative trends of the paper remain stable across seeds. In particular, \jplagDiv{} consistently produces substantially higher JPlag diversity than the other RLVR objectives while maintaining strong repeated-sampling executable performance. Although RLVR objectives exhibit noticeable variance across runs, the anti-redundancy reward does not appear to introduce additional instability relative to the baselines in these experiments.

\begin{table*}[h]
    \centering
    \resizebox{\textwidth}{!}{\input{tables/seed_table}}
    \caption{Five-seed results for Qwen3-4B on MBPP and TACO. Values report mean and standard deviation across independent training runs. Across both datasets, \jplagDiv{} consistently maintains substantially higher JPlag diversity while remaining competitive or superior on repeated-sampling executable performance.}
    \label{tab:seed-variation}
\end{table*}

\section{Diversity-Only Training Dynamics}
\label{app:diversity-only}
\begin{figure*}[h]
    \centering
    \begin{subfigure}[t]{0.28\textwidth}
        \centering
        \includegraphics[width=\linewidth]{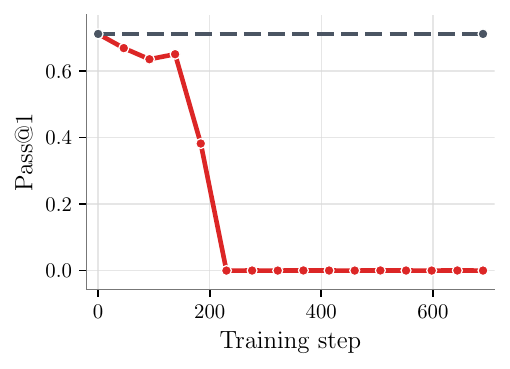}
        \caption{\passat{1}}
        \label{fig:diversity-only-pass1}
    \end{subfigure}
    \hfill
    \begin{subfigure}[t]{0.28\textwidth}
        \centering
        \includegraphics[width=\linewidth]{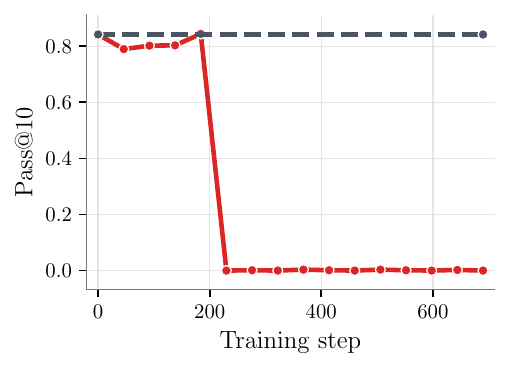}
        \caption{\passat{10}}
        \label{fig:diversity-only-pass10}
    \end{subfigure}
    \hfill
    \begin{subfigure}[t]{0.385\textwidth}
        \centering
        \includegraphics[width=\linewidth]{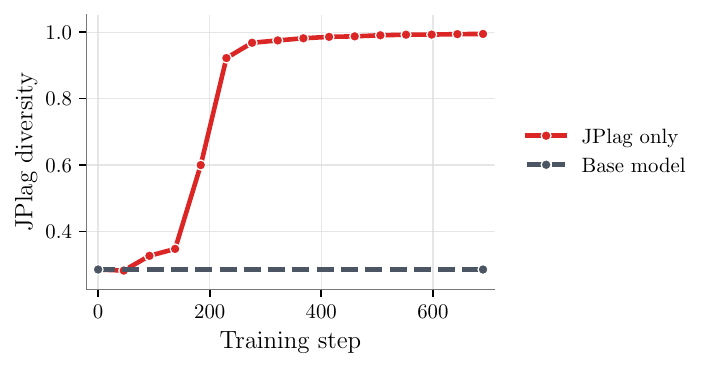}
        \caption{JPlag diversity}
        \label{fig:diversity-only-diversity}
    \end{subfigure}
    \caption{Diversity-only training for Qwen3-4B on MBPP. Optimizing the diversity reward without executable correctness increases JPlag diversity, but collapses executable performance. Effective anti-redundancy training must balance correctness and redundancy.}
    \label{fig:diversity-only-sanity}
\end{figure*}

Figure~\ref{fig:diversity-only-sanity} provides the training dynamics for the diversity-only sanity check discussed in Section~\ref{sec:ablations}. The objective increases JPlag diversity, but without executable correctness it collapses finite-budget performance.

\section{Alternative Diversity Reward Ablations}
\label{app:ablations-other-rewards}

Table~\ref{tab:other-div-mbpp} compares alternative group-level diversity rewards on Qwen3-4B and MBPP.

\paragraph{Distributional and embedding-space rewards are weak.} Entropy regularization does not improve over \baseRLVR{}, with slightly lower executable performance and nearly unchanged diversity diagnostics. The Vendi-based reward is also comparatively weak in this setting, leaving most metrics close to or below \baseRLVR{}, including the Vendi score itself. This suggests that generic distributional spreading or embedding-space diversity may not provide a sufficiently useful training signal for finite-budget code generation in this setup.

\paragraph{Surface anti-redundancy already helps.} In contrast, rewards that more directly discourage repeated generations produce substantially stronger results. The lexical 1-gram reward already yields large improvements over \baseRLVR{}, showing that even simple surface-level anti-redundancy can be an effective training signal for finite-budget code generation. The JPlag-based reward performs best overall, achieving the strongest executable performance together with the strongest redundancy diagnostics across all reported metrics.

\paragraph{JPlag better targets relevant redundancy.} These results support two conclusions. First, the gains are not specific to a single implementation of the reward: reducing repeated generations is itself useful under finite sampling budgets. Second, the stronger performance of JPlag over 1-gram diversity supports our use of a code-specific redundancy signal rather than a purely lexical one. Surface-level anti-redundancy already helps, but JPlag provides a better-aligned training signal for discouraging repeated implementations.

\begin{table}[t]
    \centering
    \small
    \input{tables/other_div_mpbb}
    \caption{Alternative group-level diversity rewards on Qwen3-4B and MBPP. \passat{1}, \passat{10}, and \passat{100} report finite-budget executable success; JDiv, 1gDiv, and Vendi report JPlag, 1-gram, and embedding-space diversity, respectively. Lexical 1-gram anti-redundancy already improves repeated-sampling performance over \baseRLVR{}, while the JPlag-based reward gives the strongest overall results and redundancy diagnostics. Entropy and Vendi-based rewards are weaker in this comparison. A star (*) marks methods significantly better than all non-diversity baselines under a paired bootstrap test over prompts (\(p < 0.05\)).}
    \label{tab:other-div-mbpp}
\end{table}

\section{Metrics Discussions}
\label{app:metrics-discussions}

\begin{figure*}[h]
    \centering

    \begin{subfigure}[t]{0.32\linewidth}
        \centering
        \includegraphics[width=\linewidth]{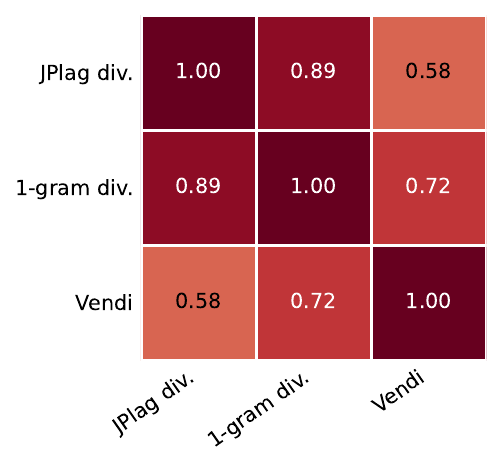}
        \caption{Code-Contest}
        \label{fig:corr-code-contest}
    \end{subfigure}
    \hfill
    \begin{subfigure}[t]{0.32\linewidth}
        \centering
        \includegraphics[width=\linewidth]{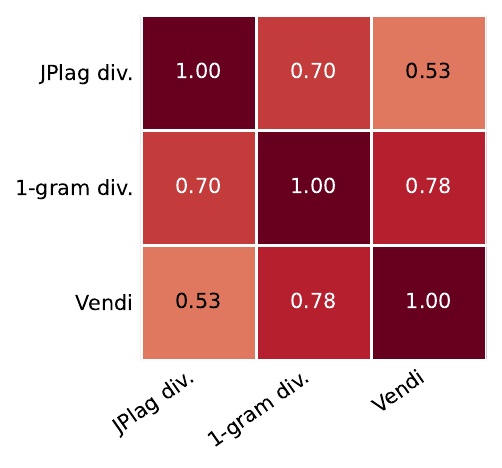}
        \caption{MBPP}
        \label{fig:corr-mbpp}
    \end{subfigure}
    \hfill
    \begin{subfigure}[t]{0.32\linewidth}
        \centering
        \includegraphics[width=\linewidth]{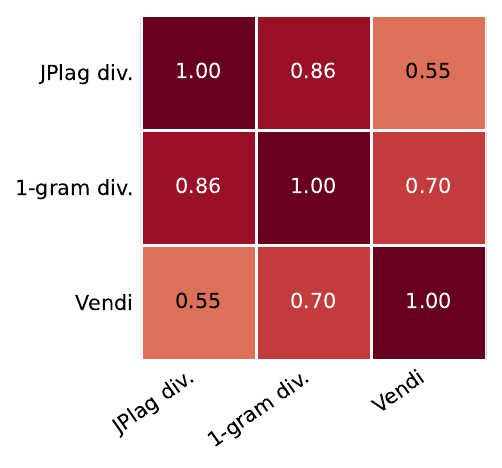}
        \caption{TACO}
        \label{fig:corr-taco}
    \end{subfigure}

    \caption{
        Correlation between diversity metrics across prompts pooled over all models and training methods for each dataset.
    }
    \label{fig:diversity-metric-correlations}
\end{figure*}

\paragraph{Relationship between diversity diagnostics.}
Figure~\ref{fig:diversity-metric-correlations} shows Pearson correlations between diversity metrics aggregated across all evaluated models and training objectives for each dataset. JPlag diversity and lexical diversity are strongly correlated across datasets, suggesting that lexical overlap often tracks implementation redundancy reasonably well. In contrast, Vendi exhibits only moderate correlation with both metrics, indicating that embedding-space dispersion captures a related but distinct notion of variation among sampled programs.

\section{JPlag Cluster and Correct-Only Analysis}
\label{app:jplag-cluster-analysis}

The main experiments report JPlag diversity as a pairwise redundancy diagnostic. Here we add a coarser cluster-based view of the sampled implementations. For each prompt, we construct a threshold graph over generated programs by connecting two programs when their JPlag similarity exceeds \(\tau=0.7\). The connected components of this graph define groups of near-duplicate implementations.

\paragraph{Cluster-based redundancy diagnostics.}
Table~\ref{tab:jplag-analysis} reports two kinds of JPlag-based coverage diagnostics. JDiv is the JPlag diversity score from Equation~\eqref{eq:jplag-diversity}, computed over all sampled programs. Eff is the effective number of JPlag clusters,
\[
    \mathrm{eff}(Y_x)
    =
    \exp\left(- \sum_r p_r \log p_r\right),
\]
where \(p_r\) is the fraction of samples assigned to cluster \(r\). Unlike the raw number of clusters, the effective cluster count is lower when most generations fall into a few dominant implementation families. It therefore distinguishes broad, balanced implementation coverage from a long tail of rare variants.

\paragraph{Cluster-based redundancy diagnostics.}
Table~\ref{tab:jplag-analysis} reports two kinds of JPlag-based diagnostics. JDiv is the pairwise JPlag diversity score from Equation~\eqref{eq:jplag-diversity}, computed over all sampled programs. Eff is the effective number of JPlag clusters,
\[
    \mathrm{eff}(Y_x)
    =
    \exp\left(- \sum_r p_r \log p_r\right),
\]
where \(p_r\) is the fraction of samples assigned to cluster \(r\). Unlike the raw number of clusters, the effective cluster count is lower when most generations fall into a few dominant near-duplicate groups. It therefore measures whether samples are distributed across multiple substantial JPlag-based clusters, rather than dominated by a single near-duplicate group with a long tail of rare variants.

\paragraph{Correct-only diagnostics.}
The columns marked with ``-c'' compute the same diagnostics after filtering to correct programs only. JDiv-c measures pairwise JPlag diversity among successful solutions, and Eff-c measures the effective number of JPlag clusters among successful solutions. This split is important because a model can increase diversity by generating many incorrect variants. The correct-only metrics test whether lower redundancy also appears within the useful part of the sampled set.

\begin{table*}[t]
    \centering
    \resizebox{\linewidth}{!}{\input{tables/jplag_analysis.tex}}
    \caption{JPlag-based redundancy and cluster diagnostics across all sampled programs and correct-only programs. JDiv is mean pairwise JPlag diversity. Eff is the effective number of JPlag clusters obtained from a similarity threshold graph with \(\tau=0.7\). The \(-c\) columns compute the corresponding metrics after filtering to correct programs only.}
    \label{tab:jplag-analysis}
\end{table*}

The table shows that \jplagDiv{} usually produces the strongest JPlag-based implementation coverage on MBPP, both over all samples and among correct solutions. For example, with Qwen3-8B on MBPP, \jplagDiv{} increases JDiv from \(0.305\) to \(0.927\) and Eff from \(18.95\) to \(90.64\); after filtering to correct programs, JDiv-c remains high at \(0.921\) and Eff-c increases from \(17.44\) to \(81.19\). This indicates that the diversity gain is not merely due to incorrect exploratory outputs: successful solutions themselves are spread across many more substantial JPlag-based implementation clusters.

The pattern is more heterogeneous on Code-Contest and TACO, where \passk-aware objectives such as \passkRLVR{} and \pkpo{} sometimes obtain the largest all-sample cluster dispersion. However, \jplagDiv{} remains competitive and is often strongest on correct-only diagnostics, particularly on TACO where it achieves the highest JDiv-c across all three models. These results suggest that direct anti-redundancy optimization reduces redundancy among successful implementations rather than simply increasing variation across all generated programs.

\begin{table*}[t]
    \centering
    \input{tables/cluster_representatives}
    \caption{Representative generations from three JPlag clusters for one MBPP prompt: ``Write a function to determine if there is a subset of the given set with sum equal to the given sum.'' The clusters correspond to recognizable implementation families, including set-based dynamic programming, boolean-table dynamic programming, and recursive search.}
    \label{tab:cluster-representatives}
\end{table*}

Table~\ref{tab:cluster-representatives} provides illustrative examples of the structure captured by the cluster analysis. In this example, JPlag groups near-duplicate generations into coherent implementation families rather than only detecting surface-level lexical overlap. The representative clusters differ in control flow and state representation, suggesting that the effective cluster count can reflect meaningful variation across recognizable programming strategies.

\section{Implementation Details}
\label{sec:implementation-details}

\paragraph{Training setup and hyperparameters.}
All RLVR methods use the same optimization hyperparameters within each dataset. We use a learning rate of \(10^{-6}\). Models are trained for \(15\) epochs on MBPP and \(6\) epochs on Code-Contest and TACO-Cobalt. Each optimization batch contains \(8\) prompts with \(32\) sampled generations per prompt for MBPP, and \(4\) prompts with \(32\) generations per prompt for Code-Contest and TACO-Cobalt. We reduce the number of prompts per batch on competitive-programming datasets to control execution cost during RLVR training.

Optimization hyperparameters were selected using \baseRLVR{} validation performance and then kept fixed across all RL objectives.  We use validation \passat{10} for checkpoint selection.

For the baselines, we follow the recommendations of \passkRLVR{} and \pkpo{} original works and set the objective parameter \(k\) to match the evaluation target, i.e., \(k=10\). For \passkRLVR{}, this additionally requires reducing the sampled group size to \(10\) generations per prompt.

For redundancy-aware objectives, we select the anti-redundancy coefficient \(\lambda_{\mathrm{div}}\) using validation \passat{10}. We search over \(\{1,2,4,6\}\) on MBPP and over \(\{0.25,0.5,1.0,1.5,2.0\}\) on Code-Contest and TACO-Cobalt. We use dataset-specific search grids because we empirically observed that the useful coefficient range varies across datasets, potentially due to differences in task difficulty, code structure, and implementation variability.

\paragraph{Datasets and splits.}
Table~\ref{tab:dataset-splits} summarizes the train, validation, and test splits used in our experiments. We use three executable code-generation benchmarks:
\begin{itemize*}
    \item \textbf{MBPP} \citep{mbpp} contains short crowd-sourced Python programming tasks designed around entry-level programming concepts, each paired with reference code and unit tests. We use the standard train/validation/test split; the public dataset is distributed under CC-BY-4.0.
    \item \textbf{Code-Contest} \citep{li2022competition} contains competitive-programming problems from online judges, with problem statements, input-output tests, and human submissions in multiple languages. We use the official validation and test sets; the public release licenses code under Apache-2.0 and non-code materials under CC-BY-4.0.
    \item \textbf{TACO-Cobalt} \citep{taco-cobalt,taco-original} is a cleaned version of TACO focused on competition-style algorithmic code generation, with public and hidden tests for each task. It does not provide an official validation split, so we reserve \(300\) training examples for validation; the public dataset card lists an MIT license.
\end{itemize*}
For all datasets, we use a simple instruction template that presents the programming task and requests a Python solution in a standardized executable format:

\begin{quote}
    \small
    \begin{verbatim}
You are an expert Python programmer, and here is your task:
{{ current_problem.content }}

Your solution should be in Python and follow this format:
```python
[your code here]
```
\end{verbatim}
\end{quote}

\begin{table}[h]
    \centering
    \small

    \begin{tabular}{lccc}
        \toprule
        Dataset      & Train Size           & Validation Size         & Test Size \\
        \midrule
        MBPP         & $374$                & $90$                    & $500$     \\
        Code-Contest & $1{,}000 / 13{,}328$ & $117$                   & $165$     \\
        TACO-Cobalt  & $1{,}000 / 5{,}853$  & $300 / 5{,}853$ (train) & $250$     \\
        \bottomrule
    \end{tabular}
    \caption{Dataset splits used in our experiments. Fractions indicate the subset size used out of the available split.}
    \label{tab:dataset-splits}
\end{table}

To control computational cost during RLVR training, we subsample \(1000\) training examples from Code-Contest and TACO-Cobalt. We use the full MBPP training split because of its smaller size and lower execution cost.

\paragraph{Models and compute.}
We run experiments with Qwen3-4B, Qwen3-8B \citep{yang2025qwen3technicalreport}, and Olmo3-7B \citep{olmo2026olmo3}, covering models with approximately \(4\), \(8\), and \(7\) billion parameters. Across all training, validation, and test-generation runs, the experiments used approximately \(15{,}000\) GPU hours on NVIDIA H100 and A100 GPUs.

\paragraph{Software stack.}
Training is implemented in PyTorch \citep{paszke2019pytorch} using Hugging Face Transformers \citep{wolf2020transformers} and TRL \citep{vonwerra2020trl}. We use vLLM \citep{kwon2023efficient} for inference.

\paragraph{Sandbox and execution environment.}
We use SandboxFusion~\citep{bytedance2025fullstackbench} as the execution backend for both training and evaluation.\footnote{\url{https://github.com/bytedance/SandboxFusion}} SandboxFusion provides isolated execution and benchmark evaluation for LLM-generated code across multiple programming datasets and programming languages. For all experiments, generated completions are executed inside the sandbox environment and evaluated against the corresponding unit tests or task checkers before computing RL rewards and diversity diagnostics. SandboxFusion is licensed under the Apache License, Version 2.0.

\paragraph{JPlag implementation details.}
We use JPlag v6.2.0~\citep{prechelt2002jplag} through a local containerized service based on the official GPL-3.0 release.\footnote{\url{https://github.com/jplag/JPlag}} Generated completions are first executed in the sandbox environment, from which we extract the submitted Python code before computing similarity scores.

For both training and evaluation, we compute pairwise JPlag similarities between generated programs for the same prompt using the Python 3 frontend with \texttt{SimilarityMetric.AVG}, \texttt{minimum\_token\_match=5}. We use the returned \texttt{averageSimilarity} value as the pairwise similarity score. JPlag clustering is disabled during similarity computation; all cluster-based diagnostics reported in the appendix are computed afterward from the resulting pairwise similarity matrix.

\paragraph{1-gram diversity metric.}
For the lexical diversity metric, we first remove comments and docstrings from each generated program. The remaining code is then tokenized using the \texttt{tiktoken} tokenizer\footnote{\url{https://github.com/openai/tiktoken}}. Pairwise 1-gram overlap is computed over the resulting token sequences, and converted into a diversity score analogously to the JPlag-based metric.

\paragraph{Checkpoint selection.}
We select checkpoints using validation \passat{10}. For MBPP, validation uses the full validation split with \(100\) sampled generations per prompt. For Code-Contest, we use the full validation split with \(50\) sampled generations per prompt to control repeated-sampling evaluation cost. For TACO-Cobalt, we use the \(300\)-example validation subset together with \(50\) sampled generations per prompt, again to control computational cost during validation. These validation settings are kept fixed across all methods.

\paragraph{Reward encoding.}
In the main text, we write verifier rewards as \(r_i = c(x,y_i)\) for readability. In the implementation, binary correctness values \(c(x,y_i) \in \{0,1\}\) are mapped to signed rewards in \(\{-1,1\}\) before applying the policy-gradient estimators. This affine transformation does not change the relative ordering of samples or the leave-one-out advantages used by the training objectives.

\paragraph{AI usage.}
We used AI assistants during the development and writing process, including for code prototyping, debugging, experiment-management scripts, and language polishing. All experimental results, analyses, and paper claims were checked by the authors.

\clearpage

\section{Completion Length Diagnostics}
\label{app:completion-lengths}

\begin{figure*}[h]
    \centering
    \begin{subfigure}[t]{0.495\textwidth}
        \centering
        \includegraphics[width=\linewidth]{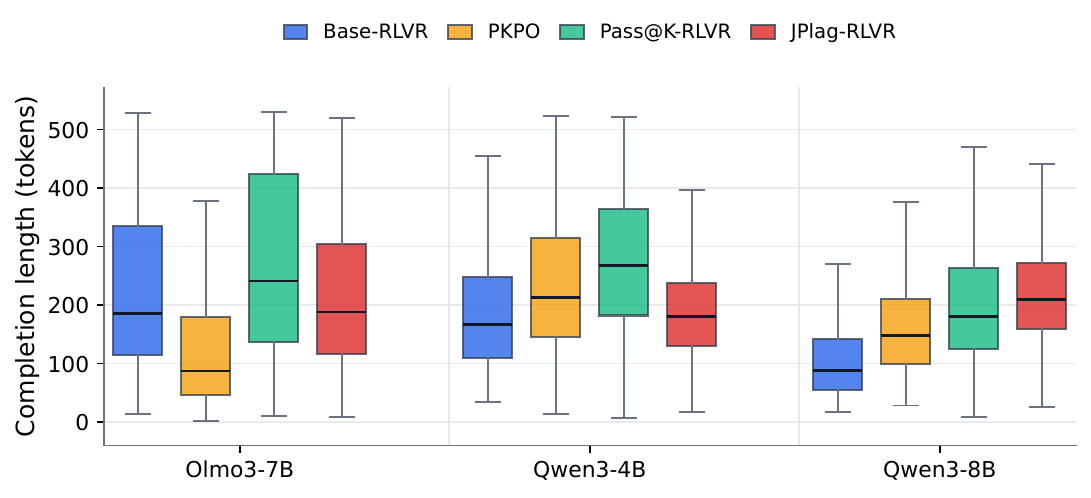}
        \caption{MBPP, raw completions}
        \label{fig:length-mbpp-raw}
    \end{subfigure}
    \hfill
    \begin{subfigure}[t]{0.495\textwidth}
        \centering
        \includegraphics[width=\linewidth]{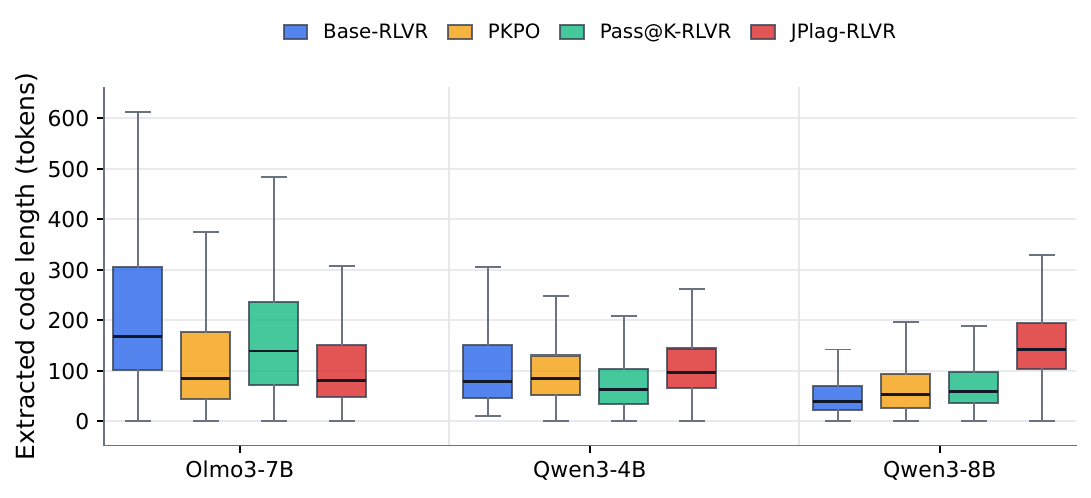}
        \caption{MBPP, extracted code}
        \label{fig:length-mbpp-code}
    \end{subfigure}

    \begin{subfigure}[t]{0.495\textwidth}
        \centering
        \includegraphics[width=\linewidth]{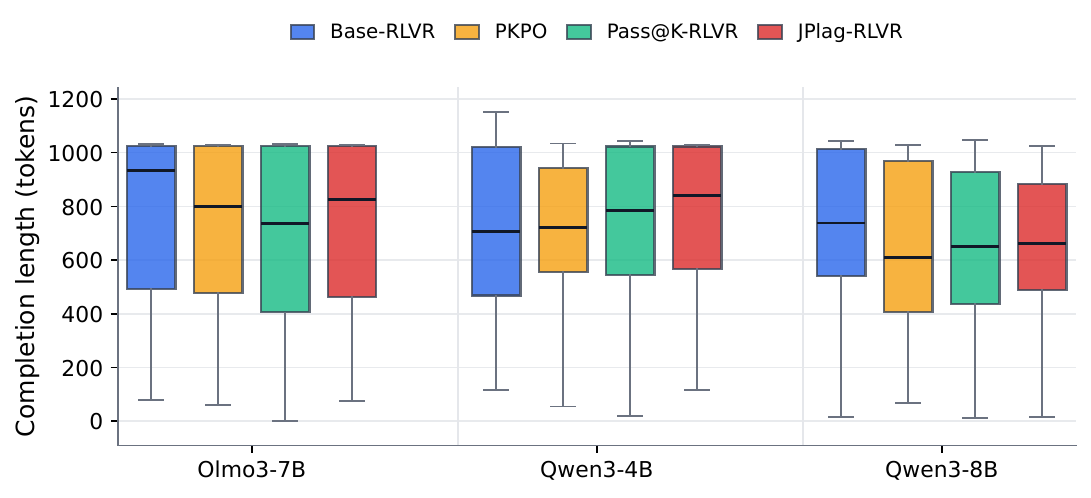}
        \caption{Code-Contest, raw completions}
        \label{fig:length-code-contest-raw}
    \end{subfigure}
    \hfill
    \begin{subfigure}[t]{0.495\textwidth}
        \centering
        \includegraphics[width=\linewidth]{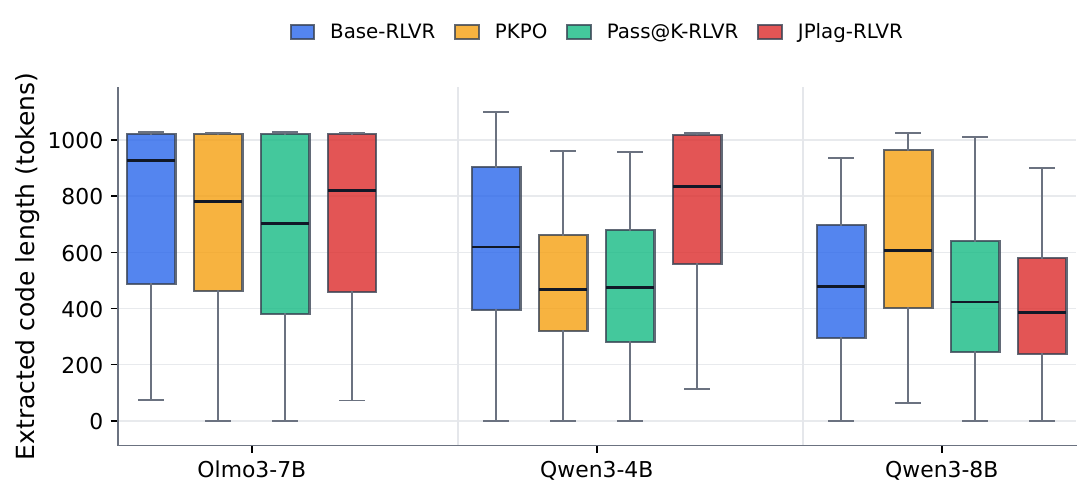}
        \caption{Code-Contest, extracted code}
        \label{fig:length-code-contest-code}
    \end{subfigure}

    \begin{subfigure}[t]{0.495\textwidth}
        \centering
        \includegraphics[width=\linewidth]{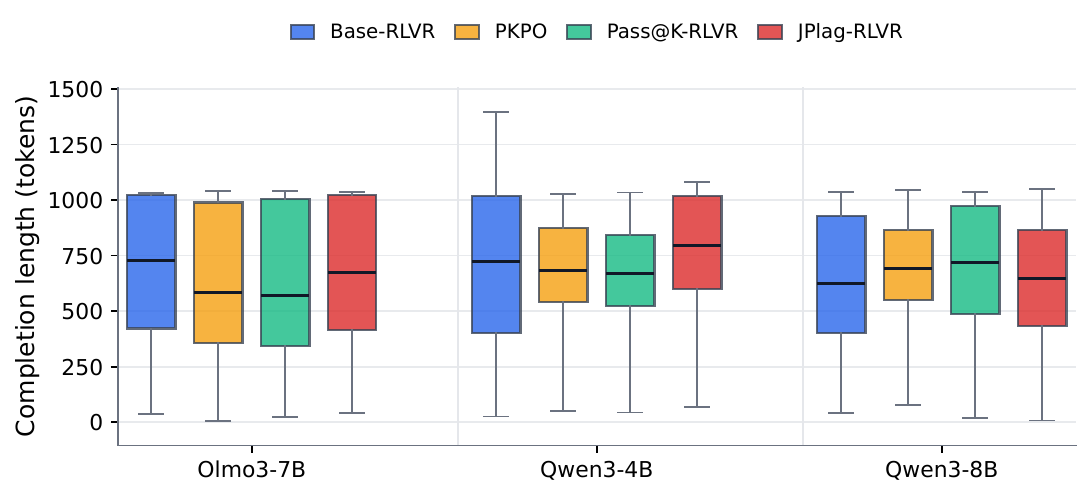}
        \caption{TACO-Cobalt, raw completions}
        \label{fig:length-taco-raw}
    \end{subfigure}
    \hfill
    \begin{subfigure}[t]{0.495\textwidth}
        \centering
        \includegraphics[width=\linewidth]{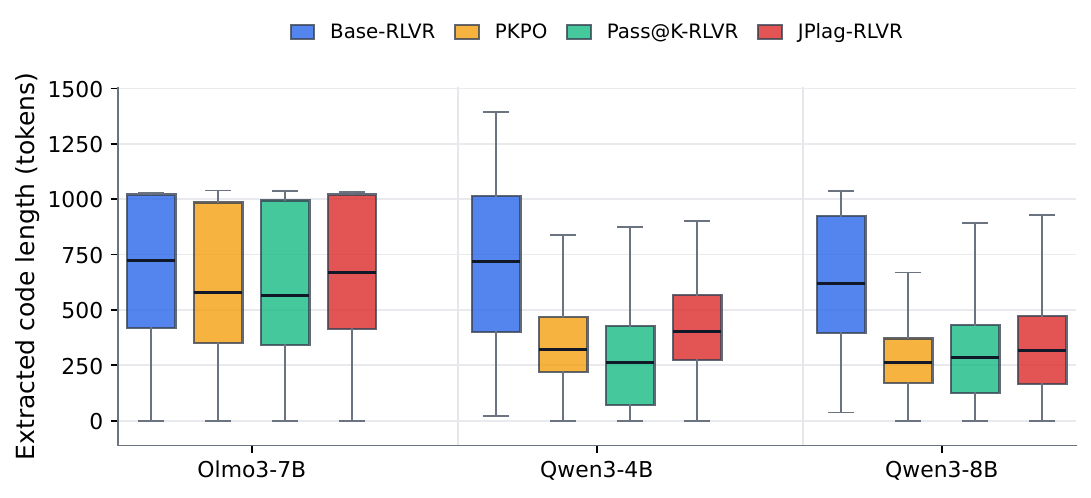}
        \caption{TACO-Cobalt, extracted code}
        \label{fig:length-taco-code}
    \end{subfigure}
    \caption{Completion-length diagnostics across datasets and training methods. Each row corresponds to one dataset, with raw completion lengths on the left and lengths after extracting executable code on the right. Diversity improvements are not accompanied by systematic inflation in either raw completions or extracted code.}
    \label{fig:completion-lengths}
\end{figure*}

Figure~\ref{fig:completion-lengths} compares raw completion lengths and extracted executable-code lengths across methods. These diagnostics address the possibility that diversity rewards reduce measured redundancy through superficial reward hacking, such as producing longer conversational completions, chain-of-thought padding, or markdown artifacts. Across MBPP, Code-Contest, and TACO-Cobalt, length distributions remain broadly comparable, suggesting that the diversity gains in the main experiments are not explained by trivial length inflation.

\clearpage

\section{Additional Baseline Training Trajectories}
\label{app:rlvr-trajectories}
\input{appendix_training_trajectories.tex}

\end{document}

%% file: tables/qualitative_jplag_examples.tex
\begin{tabular}{@{}p{0.07\linewidth}p{0.43\linewidth}p{0.43\linewidth}@{}}
    \toprule
    Scores & Generation A & Generation B \\
    \midrule

    \begin{minipage}[c]{\linewidth}
        JPlag: 1.00\par
        1-gram: 0.45
    \end{minipage}
           &
    \begin{minipage}[c]{\linewidth}
        \begin{verbatim}
def max_val(lst):
    numbers = [x for x in lst
               if isinstance(x, int)]
    return max(numbers)
        \end{verbatim}
    \end{minipage}
           &
    \begin{minipage}[c]{\linewidth}
        \begin{verbatim}
def max_val(het_list):
    numbers = [item for item in het_list
               if isinstance(item, int)]
    return max(numbers)
        \end{verbatim}
    \end{minipage}
    \\
    \midrule
    \begin{minipage}[c]{\linewidth}
        JPlag: 0.68\par
        1-gram: 0.56
    \end{minipage}
           &
    \begin{minipage}[c]{\linewidth}
        \begin{verbatim}
def max_val(het_list):
    max_value = float('-inf')
    for item in het_list:
        if (isinstance(item, int) or (isinstance(item, str)
                and item.isdigit())):
            num = int(item)
            max_value = num if num > max_value else max_value
    return max_value
        \end{verbatim}
    \end{minipage}
           &
    \begin{minipage}[c]{\linewidth}
        \begin{verbatim}
def max_val(het_list):
    numbers = [
        int(item) for item in het_list
        if (isinstance(item, int) or (isinstance(item, str)
             and item.isdigit()))
    ]
    return max(numbers) if numbers else None
        \end{verbatim}
    \end{minipage}
    \\
    \bottomrule
\end{tabular}

%% file: tables/baselines_comparison/aggregated_results.tex
\begingroup
\setlength{\tabcolsep}{3pt}
\begin{tabular}{cclllcc}
    \toprule
                  &                & \multicolumn{3}{c}{Executable}     & \multicolumn{2}{c}{Redundancy}                                                                                                                    \\
    \cmidrule(lr){3-5}
    \cmidrule(lr){6-7}
    Meth.         & Change         & p@1                                & p@10                               & p@100                              & JDiv                               & Vendi                              \\
    \midrule
    \baseRLVR     & $\uparrow$\%   & \textcolor{green!50!black}{45.5\%} & \textcolor{green!50!black}{37.6\%} & \textcolor{green!50!black}{24.5\%} & 36.1\%                             & 33.3\%                             \\
                  & $\downarrow$\% & 13.5\%                             & 10.6\%                             & 5.6\%                              & \textcolor{red!70!black}{57.2\%}   & \textcolor{red!70!black}{65.3\%}   \\
                  & $\Delta$       & \textcolor{green!50!black}{9.5}    & \textcolor{green!50!black}{11.5}   & \textcolor{green!50!black}{10.6}   & \textcolor{red!70!black}{-0.046}   & \textcolor{red!70!black}{-0.53}    \\
    \midrule
    \passkRLVR    & $\uparrow$\%   & \textcolor{green!50!black}{38.9\%} & \textcolor{green!50!black}{35.4\%} & \textcolor{green!50!black}{25.8\%} & \textcolor{green!50!black}{66.9\%} & \textcolor{green!50!black}{73.3\%} \\
                  & $\downarrow$\% & 29.6\%                             & 19.7\%                             & 5.5\%                              & 32.4\%                             & 26.6\%                             \\
                  & $\Delta$       & \textcolor{green!50!black}{4.5}    & \textcolor{green!50!black}{11.3}   & \textcolor{green!50!black}{12.4}   & \textcolor{green!50!black}{0.123}  & \textcolor{green!50!black}{0.32}   \\
    \midrule
    \pkpo         & $\uparrow$\%   & \textcolor{green!50!black}{45.7\%} & \textcolor{green!50!black}{42.4\%} & \textcolor{green!50!black}{30.0\%} & \textcolor{green!50!black}{60.0\%} & \textcolor{green!50!black}{57.6\%} \\
                  & $\downarrow$\% & 26.7\%                             & 15.7\%                             & 3.6\%                              & 38.9\%                             & 42.4\%                             \\
                  & $\Delta$       & \textcolor{green!50!black}{7.2}    & \textcolor{green!50!black}{14.9}   & \textcolor{green!50!black}{15.5}   & \textcolor{green!50!black}{0.083}  & \textcolor{red!70!black}{-0.13}    \\
    \midrule
    \jplagDiv     & $\uparrow$\%   & \textcolor{green!50!black}{48.4\%} & \textcolor{green!50!black}{46.2\%} & \textcolor{green!50!black}{32.3\%} & \textcolor{green!50!black}{77.4\%} & \textcolor{green!50!black}{73.8\%} \\
    \emph{(Our) } & $\downarrow$\% & 28.7\%                             & 12.1\%                             & 2.2\%                              & 22.6\%                             & 26.2\%                             \\
                  & $\Delta$       & \textcolor{green!50!black}{16.4}   & \textcolor{green!50!black}{19.7}   & \textcolor{green!50!black}{17.3}   & \textcolor{green!50!black}{0.298}  & \textcolor{green!50!black}{0.33}   \\
    \bottomrule
\end{tabular}
\endgroup

%% file: tables/main_results.tex
\begingroup
\setlength{\tabcolsep}{1.8pt}
\renewcommand{\arraystretch}{1.05}
\scriptsize
\begin{tabular}{lcccccccccccccccccc}
        \toprule
                               & \multicolumn{6}{c}{\textbf{MBPP}} & \multicolumn{6}{c}{\textbf{Code-Contest}} & \multicolumn{6}{c}{\textbf{TACO}}                                                                                                                                                                                                                                                                                                                         \\
        \cmidrule(lr){2-7} \cmidrule(lr){8-13} \cmidrule(lr){14-19}
                               & p@1                               & p@10                                      & p@100                             & JDiv.                & 1gDiv.               & Vendi               & p@1                 & p@10                & p@100         & JDiv.          & 1gDiv.         & Vendi         & p@1                 & p@10                & p@100               & JDiv.                & 1gDiv.               & Vendi               \\
        \midrule
        \textsc{Qwen3-4b}                                                                                                                                                                                                                                                                                                                                                                                                                                                  \\
        \midrule
        \baseRLVR              & 72.4                              & 82.4                                      & 86.8                              & 0.301                & 0.153                & 1.23                & 16.8                & 31.5                & 39.6          & 0.396          & 0.375          & 2.32          & 24.0                & 39.8                & 48.9                & 0.357                & 0.363                & 2.13                \\
        \passkRLVR             & 73.6                              & 87.8                                      & 92.9                              & 0.389                & 0.218                & 1.30                & 12.1                & 29.3                & 41.4          & \textbf{0.572} & \textbf{0.485} & \textbf{3.29} & 18.4                & 39.9                & 55.4                & 0.503                & 0.475                & 3.04                \\
        \pkpo                  & 72.5                              & 86.5                                      & 92.2                              & 0.462                & 0.279                & 1.47                & 13.4                & 31.7                & 46.0          & 0.410          & 0.350          & 2.40          & 19.5                & 42.5                & 56.0                & 0.497                & 0.430                & 2.62                \\
        \arrayrulecolor{black!35}\specialrule{0.08pt}{0.35ex}{0.35ex}\arrayrulecolor{black}
        \jplagDiv \emph{(Our)} & \textbf{93.2$^{*}$}               & \textbf{98.9$^{*}$}                       & \textbf{99.3$^{*}$}               & \textbf{0.822$^{*}$} & \textbf{0.446$^{*}$} & \textbf{1.82$^{*}$} & \textbf{17.8}       & \textbf{35.3$^{*}$} & \textbf{46.2} & 0.523          & 0.460          & 2.92          & \textbf{27.4$^{*}$} & \textbf{46.7$^{*}$} & \textbf{60.2$^{*}$} & \textbf{0.640$^{*}$} & \textbf{0.526$^{*}$} & \textbf{3.29$^{*}$} \\
        \midrule
        \textsc{Qwen3-8b}                                                                                                                                                                                                                                                                                                                                                                                                                                                  \\
        \midrule
        \baseRLVR              & 79.1                              & 88.0                                      & 90.8                              & 0.305                & 0.143                & 1.24                & 17.4                & 33.5                & 45.6          & 0.370          & 0.348          & 2.37          & 29.2                & 48.2                & 58.5                & 0.337                & 0.362                & 2.33                \\
        \passkRLVR             & 74.5                              & 89.5                                      & 93.6                              & 0.510                & 0.326                & 1.77                & 13.5                & 33.9                & 47.6          & \textbf{0.495} & \textbf{0.455} & \textbf{3.39} & 15.8                & 37.7                & 54.1                & 0.522                & 0.478                & \textbf{3.89}       \\
        \pkpo                  & 79.3                              & 97.1                                      & 99.0                              & 0.583                & 0.342                & 1.78                & 15.0                & 32.0                & 45.6          & 0.410          & 0.366          & 2.43          & 25.5                & 51.0                & 64.7                & 0.501                & 0.412                & 2.95                \\
        \arrayrulecolor{black!35}\specialrule{0.08pt}{0.35ex}{0.35ex}\arrayrulecolor{black}
        \jplagDiv \emph{(Our)} & \textbf{90.0$^{*}$}               & \textbf{98.5$^{*}$}                       & \textbf{99.2$^{*}$}               & \textbf{0.927$^{*}$} & \textbf{0.598$^{*}$} & \textbf{2.71$^{*}$} & \textbf{20.3$^{*}$} & \textbf{35.8}       & \textbf{47.7} & 0.466          & 0.411          & 2.39          & \textbf{30.3}       & \textbf{55.5$^{*}$} & \textbf{68.7$^{*}$} & \textbf{0.696$^{*}$} & \textbf{0.548$^{*}$} & 3.34                \\
        \midrule
        \textsc{Olmo3-7b}                                                                                                                                                                                                                                                                                                                                                                                                                                                  \\
        \midrule
        \baseRLVR              & 64.8                              & 84.1                                      & 91.9                              & 0.555                & 0.395                & 1.81                & \textbf{14.5}       & 29.0                & 41.2          & 0.465          & 0.438          & 3.20          & \textbf{26.3}       & 45.5                & 58.0                & 0.426                & 0.435                & 2.75                \\
        \passkRLVR             & 60.9                              & 83.7                                      & 92.5                              & 0.632                & 0.446                & 2.21                & 7.4                 & 20.1                & 35.0          & \textbf{0.536} & \textbf{0.554} & \textbf{4.99} & 14.7                & 37.4                & 55.1                & 0.544                & 0.505                & \textbf{3.59}       \\
        \pkpo                  & 64.7                              & 86.7                                      & 94.0                              & 0.569                & 0.373                & 1.85                & 11.9                & \textbf{30.5}       & \textbf{45.0} & 0.516          & 0.496          & 3.58          & 19.7                & 44.1                & \textbf{59.3}       & \textbf{0.573}       & \textbf{0.552}       & 3.47                \\
        \arrayrulecolor{black!35}\specialrule{0.08pt}{0.35ex}{0.35ex}\arrayrulecolor{black}
        \jplagDiv \emph{(Our)} & \textbf{71.6$^{*}$}               & \textbf{95.0$^{*}$}                       & \textbf{98.2$^{*}$}               & \textbf{0.872$^{*}$} & \textbf{0.598$^{*}$} & \textbf{2.71$^{*}$} & 14.3                & 30.1                & 41.2          & 0.495          & 0.474          & 3.33          & 26.0                & \textbf{46.5}       & 58.9                & 0.549                & 0.503                & 3.15                \\
        \bottomrule
\end{tabular}
\endgroup

%% file: tables/temperature_mpbb.tex
\begingroup
\setlength{\tabcolsep}{3pt}
\begin{tabular}{lllllll}
    \toprule
    Method    & $T$   & p@1  & p@10 & p@100 & JDiv  & Vendi \\
    \midrule
    \baseRLVR & $1$   & 75.0 & 85.4 & 90.0  & 0.297 & 1.21  \\
    \baseRLVR & $1.5$ & 74.4 & 86.7 & 92.0  & 0.356 & 1.29  \\
    \baseRLVR & $2$   & 70.9 & 87.4 & 92.4  & 0.417 & 1.47  \\
    \arrayrulecolor{black!35}\specialrule{0.08pt}{0.35ex}{0.35ex}\arrayrulecolor{black}
    \jplagDiv & $1$   & 93.2 & 98.9 & 99.3  & 0.822 & 1.82  \\
    \bottomrule
\end{tabular}
\endgroup

%% file: tables/seed_table.tex
\begingroup
\footnotesize
\setlength{\tabcolsep}{2.5pt}
\begin{tabular}{lcccccccccccc}
    \toprule
    Method                 & \multicolumn{6}{c}{\textbf{MBPP}} & \multicolumn{6}{c}{\textbf{TACO}}                                                                                                                                                                                                                                                                                   \\
    \cmidrule(lr){2-7} \cmidrule(lr){8-13}
                           & p@1                               & p@10                              & p@100                   & JDiv                       & 1gDiv                      & Vendi                    & p@1                     & p@10                    & p@100                   & JDiv                       & 1gDiv                      & Vendi                    \\
    \midrule
    \multicolumn{13}{l}{\textsc{qwen3-4b}}                                                                                                                                                                                                                                                                                                                                           \\
    \midrule
    \baseRLVR              & 72.3 $\pm$ 1.6                    & 82.7 $\pm$ 0.5                    & 87.3 $\pm$ 0.5          & 0.306 $\pm$ 0.024          & 0.158 $\pm$ 0.017          & 1.23 $\pm$ 0.05          & 25.2 $\pm$ 1.4          & 40.8 $\pm$ 1.2          & 51.2 $\pm$ 2.0          & 0.368 $\pm$ 0.049          & 0.363 $\pm$ 0.035          & 2.31 $\pm$ 0.15          \\
    \passkRLVR             & 71.6 $\pm$ 3.2                    & 87.6 $\pm$ 3.6                    & 93.2 $\pm$ 3.0          & 0.466 $\pm$ 0.065          & 0.285 $\pm$ 0.062          & 1.58 $\pm$ 0.20          & 15.7 $\pm$ 2.8          & 35.1 $\pm$ 4.4          & 49.7 $\pm$ 4.6          & 0.470 $\pm$ 0.034          & 0.445 $\pm$ 0.043          & 3.23 $\pm$ 0.36          \\
    \pkpo                  & 79.0 $\pm$ 8.0                    & 92.4 $\pm$ 6.0                    & 95.7 $\pm$ 3.2          & 0.474 $\pm$ 0.055          & 0.335 $\pm$ 0.069          & 1.74 $\pm$ 0.24          & 20.6 $\pm$ 1.4          & 42.3 $\pm$ 1.1          & 56.4 $\pm$ 1.7          & 0.448 $\pm$ 0.043          & 0.399 $\pm$ 0.020          & 2.68 $\pm$ 0.10          \\
    \jplagDiv \emph{(Our)} & \textbf{89.8} $\pm$ 7.5           & \textbf{97.3} $\pm$ 3.7           & \textbf{98.5} $\pm$ 1.8 & \textbf{0.769} $\pm$ 0.049 & \textbf{0.454} $\pm$ 0.028 & \textbf{1.94} $\pm$ 0.18 & \textbf{26.5} $\pm$ 2.4 & \textbf{46.7} $\pm$ 1.4 & \textbf{59.0} $\pm$ 1.3 & \textbf{0.631} $\pm$ 0.024 & \textbf{0.515} $\pm$ 0.019 & \textbf{3.29} $\pm$ 0.17 \\
    \bottomrule
\end{tabular}
\endgroup

%% file: tables/other_div_mpbb.tex
\begingroup
\setlength{\tabcolsep}{1.8pt}
\renewcommand{\arraystretch}{1.05}
\begin{tabular}{lcccccc}
    \toprule

                           & p@1                 & p@10                & p@100         & JDiv.                & 1gDiv.               & Vendi         \\
    \midrule
    \baseRLVR              & 75.1                & 85.3                & 90.5          & 0.301                & 0.167                & 1.21          \\
    Entropy                & 74.3                & 82.8                & 87.6          & 0.302                & 0.155                & 1.20          \\
    \vendiDiv              & 76.3                & 85.4                & 90.0          & 0.285                & 0.156                & 1.19          \\
    1-gram-div             & 86.5                & 96.3                & 98.4          & 0.457                & 0.269                & 1.52          \\
    \arrayrulecolor{black!35}\specialrule{0.08pt}{0.35ex}{0.35ex}\arrayrulecolor{black}
    \jplagDiv \emph{(Our)} & \textbf{93.2$^{*}$} & \textbf{98.9$^{*}$} & \textbf{99.3} & \textbf{0.822$^{*}$} & \textbf{0.446$^{*}$} & \textbf{1.82} \\
    \bottomrule
\end{tabular}
\endgroup

%% file: tables/jplag_analysis.tex
\begingroup
\setlength{\tabcolsep}{2.2pt}
\renewcommand{\arraystretch}{1.05}
\scriptsize
\begin{tabular}{lccccccccccccccc}
    \toprule
                           & \multicolumn{5}{c}{\textbf{MBPP}} & \multicolumn{5}{c}{\textbf{Code-Contest}} & \multicolumn{5}{c}{\textbf{TACO}}                                                                                                                                                                                                         \\
    \cmidrule(lr){2-6} \cmidrule(lr){7-11} \cmidrule(lr){12-16}
                           & p@10                              & JDiv                                      & Eff                               & JDiv-c         & Eff-c          & p@10          & JDiv           & Eff            & JDiv-c         & Eff-c         & p@10          & JDiv           & Eff            & JDiv-c         & Eff-c         \\
    \midrule
    \textsc{Qwen3-4b}                                                                                                                                                                                                                                                                                                                                  \\
    \midrule
    \baseRLVR              & 82.4                              & 0.301                                     & 16.95                             & 0.275          & 15.35          & 31.5          & 0.396          & 2.67           & 0.315          & 2.00          & 39.8          & 0.357          & 4.62           & 0.274          & 1.72          \\
    \passkRLVR             & 87.8                              & 0.389                                     & 16.81                             & 0.355          & 13.94          & 29.3          & \textbf{0.572} & \textbf{47.59} & 0.411          & \textbf{3.27} & 39.9          & 0.503          & 22.50          & 0.397          & 2.98          \\
    \pkpo                  & 86.5                              & 0.462                                     & 14.88                             & 0.432          & 12.93          & 31.7          & 0.410          & 7.57           & 0.312          & 1.84          & 42.5          & 0.497          & 12.69          & 0.378          & 2.69          \\
    \jplagDiv \emph{(Our)} & \textbf{98.9}                     & \textbf{0.822}                            & \textbf{19.23}                    & \textbf{0.811} & \textbf{18.73} & \textbf{35.3} & 0.523          & 11.54          & \textbf{0.420} & 2.93          & \textbf{46.7} & \textbf{0.640} & \textbf{37.95} & \textbf{0.549} & \textbf{6.04} \\
    \midrule
    \textsc{Qwen3-8b}                                                                                                                                                                                                                                                                                                                                  \\
    \midrule
    \baseRLVR              & 88.0                              & 0.305                                     & 18.95                             & 0.275          & 17.44          & 33.5          & 0.370          & 2.55           & 0.279          & 1.59          & 48.2          & 0.337          & 2.81           & 0.263          & 1.69          \\
    \passkRLVR             & 89.5                              & 0.510                                     & 12.19                             & 0.477          & 10.94          & 33.9          & \textbf{0.495} & \textbf{15.31} & 0.378          & 2.30          & 37.7          & 0.522          & 18.25          & 0.377          & 2.61          \\
    \pkpo                  & 97.1                              & 0.583                                     & 14.81                             & 0.555          & 12.28          & 32.0          & 0.410          & 8.16           & 0.317          & 1.83          & 51.0          & 0.501          & 18.09          & 0.395          & 3.13          \\
    \jplagDiv \emph{(Our)} & \textbf{98.5}                     & \textbf{0.927}                            & \textbf{90.64}                    & \textbf{0.921} & \textbf{81.19} & \textbf{35.8} & 0.466          & 8.51           & \textbf{0.397} & \textbf{2.34} & \textbf{55.5} & \textbf{0.696} & \textbf{35.21} & \textbf{0.623} & \textbf{8.34} \\
    \midrule
    \textsc{Olmo3-7b}                                                                                                                                                                                                                                                                                                                                  \\
    \midrule
    \baseRLVR              & 84.1                              & 0.555                                     & 22.27                             & 0.502          & 13.16          & 29.0          & 0.465          & 5.68           & 0.335          & 1.76          & 45.5          & 0.426          & 8.63           & 0.326          & 2.20          \\
    \passkRLVR             & 83.7                              & 0.632                                     & 20.88                             & 0.585          & 12.94          & 20.1          & \textbf{0.536} & 11.66          & \textbf{0.433} & \textbf{2.86} & 37.4          & 0.544          & 24.64          & 0.411          & 3.56          \\
    \pkpo                  & 86.7                              & 0.569                                     & 20.11                             & 0.523          & 12.71          & \textbf{30.5} & 0.516          & \textbf{19.30} & 0.403          & 2.51          & 44.1          & \textbf{0.573} & \textbf{28.37} & 0.451          & \textbf{4.01} \\
    \jplagDiv \emph{(Our)} & \textbf{95.0}                     & \textbf{0.872}                            & \textbf{28.73}                    & \textbf{0.842} & \textbf{18.59} & 30.1          & 0.495          & 12.89          & 0.382          & 2.06          & \textbf{46.5} & 0.549          & 11.95          & \textbf{0.452} & 2.83          \\
    \bottomrule
\end{tabular}
\endgroup

%% file: tables/cluster_representatives.tex
\begingroup
\setlength{\tabcolsep}{1pt}
\scriptsize
\begin{tabular}{@{}p{0.27\linewidth}p{0.27\linewidth}p{0.45\linewidth}@{}}
    \toprule
    Cluster 1 & Cluster 2 & Cluster 3 \\
    \midrule
    \begin{minipage}[t]{\linewidth}
        \begin{verbatim}
def is_subset_sum(
    arr, n, target):
    achievable = set()
    achievable.add(0)

    for num in arr:
        new_sums = set()
        for s in achievable:
            new_sums.add(s + num)
        achievable.update(
            new_sums)

    return target in achievable
        \end{verbatim}
    \end{minipage}
              &
    \begin{minipage}[t]{\linewidth}
        \begin{verbatim}
def is_subset_sum(
    nums, n, target):
    dp = [False] * (
        target + 1)
    dp[0] = True

    for num in nums:
        for i in range(target,
                       num - 1, -1):
            if dp[i - num]:
                dp[i] = True

    return dp[target]
        \end{verbatim}
    \end{minipage}
              &
    \begin{minipage}[t]{\linewidth}
        \begin{verbatim}
def is_subset_sum(
    nums, start, target):
    if target == 0:
        return True
    if start == len(nums):
        return False
    same_prev = (
        start > 0 and
        nums[start] == nums[start - 1]
    )
    if same_prev:
        prev_fails = not is_subset_sum(
            nums, start - 1, target)
        if prev_fails:
            return is_subset_sum(
                nums, start + 1, target)
    remaining = target - nums[start]
    if is_subset_sum(
        nums, start + 1, remaining):
        return True
    return is_subset_sum(
        nums, start + 1, target)
        \end{verbatim}
    \end{minipage}
    \\
    \bottomrule
\end{tabular}
\endgroup

%% file: appendix_training_trajectories.tex
\begin{figure*}[h]
    \centering
    \begin{subfigure}[h]{0.28\textwidth}
        \centering
        \includegraphics[width=\linewidth]{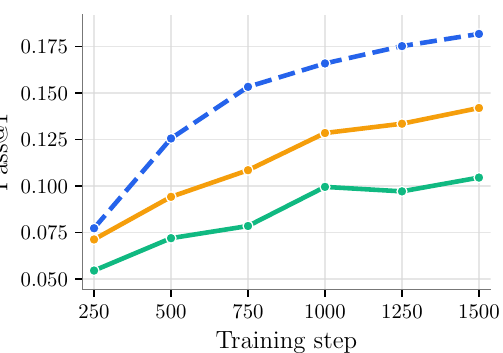}
        \caption{\passat{1}}
        \label{fig:baseline-dynamics-olmo-code-contest-pass1}
    \end{subfigure}
    \hfill
    \begin{subfigure}[h]{0.28\textwidth}
        \centering
        \includegraphics[width=\linewidth]{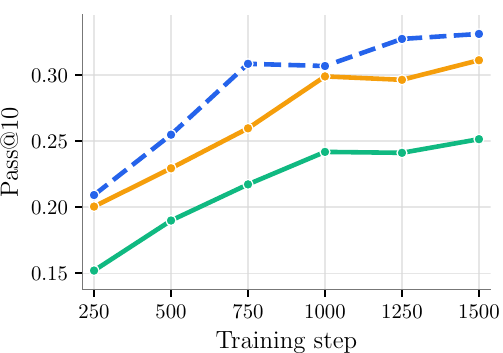}
        \caption{\passat{10}}
        \label{fig:baseline-dynamics-olmo-code-contest-pass10}
    \end{subfigure}
    \hfill
    \begin{subfigure}[h]{0.385\textwidth}
        \centering
        \includegraphics[width=\linewidth]{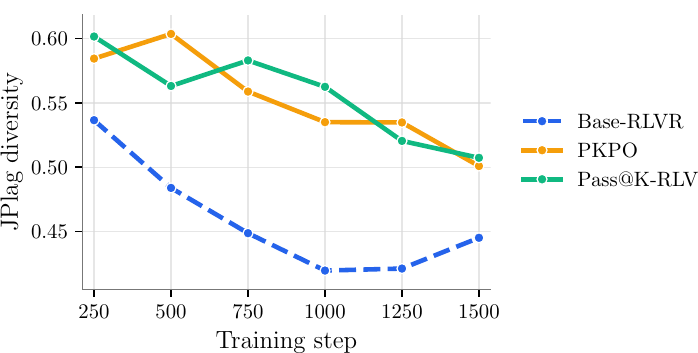}
        \caption{JPlag diversity}
        \label{fig:baseline-dynamics-olmo-code-contest-diversity}
    \end{subfigure}
    \caption{Baseline training dynamics for Olmo3-7B on Code-Contest.}
    \label{fig:baseline-dynamics-olmo-code-contest}
\end{figure*}
\begin{figure*}[h]
    \centering
    \begin{subfigure}[h]{0.28\textwidth}
        \centering
        \includegraphics[width=\linewidth]{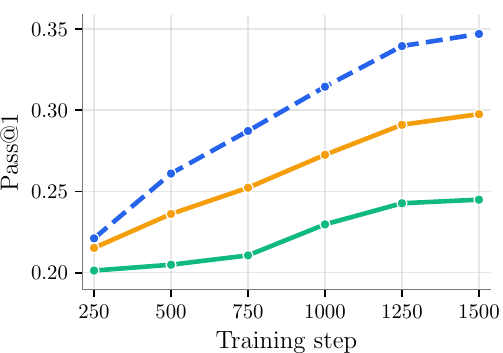}
        \caption{\passat{1}}
        \label{fig:baseline-dynamics-olmo-taco-pass1}
    \end{subfigure}
    \hfill
    \begin{subfigure}[h]{0.28\textwidth}
        \centering
        \includegraphics[width=\linewidth]{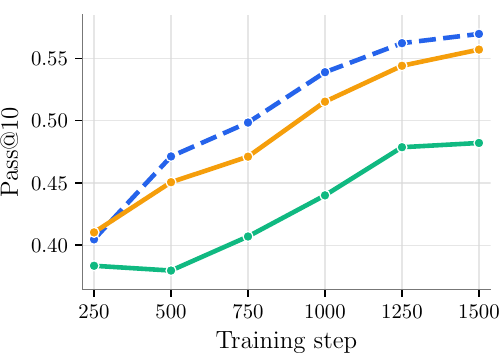}
        \caption{\passat{10}}
        \label{fig:baseline-dynamics-olmo-taco-pass10}
    \end{subfigure}
    \hfill
    \begin{subfigure}[h]{0.385\textwidth}
        \centering
        \includegraphics[width=\linewidth]{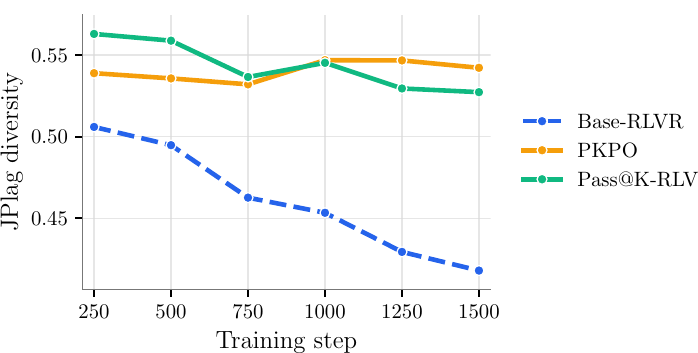}
        \caption{JPlag diversity}
        \label{fig:baseline-dynamics-olmo-taco-diversity}
    \end{subfigure}
    \caption{Baseline training dynamics for Olmo3-7B on TACO-Cobalt.}
    \label{fig:baseline-dynamics-olmo-taco}
\end{figure*}
\begin{figure*}[h]
    \centering
    \begin{subfigure}[h]{0.28\textwidth}
        \centering
        \includegraphics[width=\linewidth]{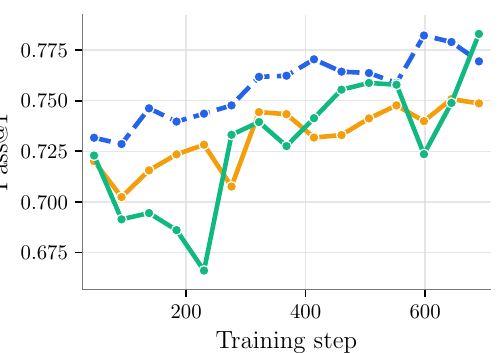}
        \caption{\passat{1}}
        \label{fig:baseline-dynamics-qwen4-mbpp-pass1}
    \end{subfigure}
    \hfill
    \begin{subfigure}[h]{0.28\textwidth}
        \centering
        \includegraphics[width=\linewidth]{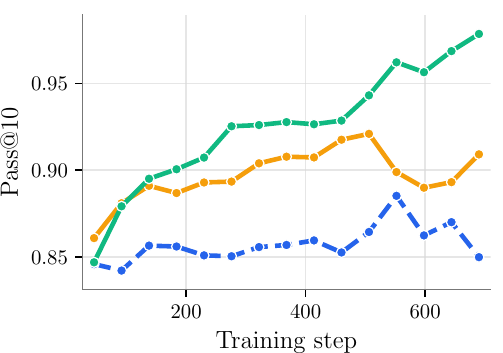}
        \caption{\passat{10}}
        \label{fig:baseline-dynamics-qwen4-mbpp-pass10}
    \end{subfigure}
    \hfill
    \begin{subfigure}[h]{0.385\textwidth}
        \centering
        \includegraphics[width=\linewidth]{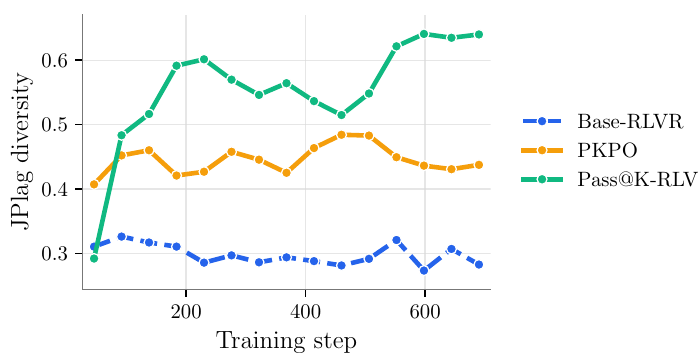}
        \caption{JPlag diversity}
        \label{fig:baseline-dynamics-qwen4-mbpp-diversity}
    \end{subfigure}
    \caption{Baseline training dynamics for Qwen3-4B on MBPP.}
    \label{fig:baseline-dynamics-qwen4-mbpp}
\end{figure*}
\begin{figure*}[h]
    \centering
    \begin{subfigure}[h]{0.28\textwidth}
        \centering
        \includegraphics[width=\linewidth]{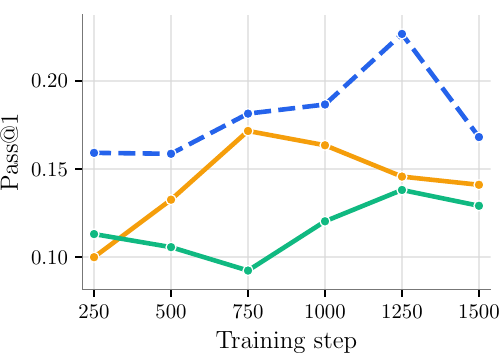}
        \caption{\passat{1}}
        \label{fig:baseline-dynamics-qwen4-code-contest-pass1}
    \end{subfigure}
    \hfill
    \begin{subfigure}[h]{0.28\textwidth}
        \centering
        \includegraphics[width=\linewidth]{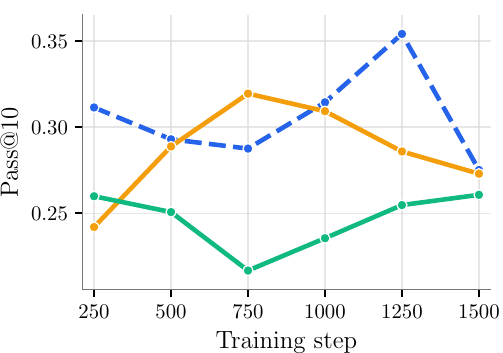}
        \caption{\passat{10}}
        \label{fig:baseline-dynamics-qwen4-code-contest-pass10}
    \end{subfigure}
    \hfill
    \begin{subfigure}[h]{0.385\textwidth}
        \centering
        \includegraphics[width=\linewidth]{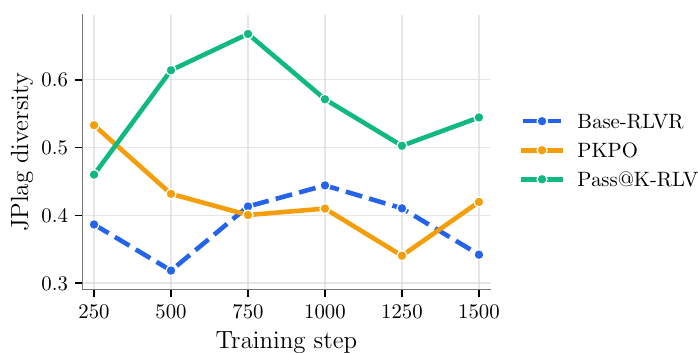}
        \caption{JPlag diversity}
        \label{fig:baseline-dynamics-qwen4-code-contest-diversity}
    \end{subfigure}
    \caption{Baseline training dynamics for Qwen3-4B on Code-Contest.}
    \label{fig:baseline-dynamics-qwen4-code-contest}
\end{figure*}
\begin{figure*}[h]
    \centering
    \begin{subfigure}[h]{0.28\textwidth}
        \centering
        \includegraphics[width=\linewidth]{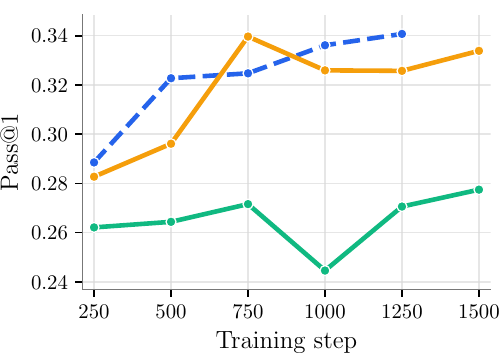}
        \caption{\passat{1}}
        \label{fig:baseline-dynamics-qwen4-taco-pass1}
    \end{subfigure}
    \hfill
    \begin{subfigure}[h]{0.28\textwidth}
        \centering
        \includegraphics[width=\linewidth]{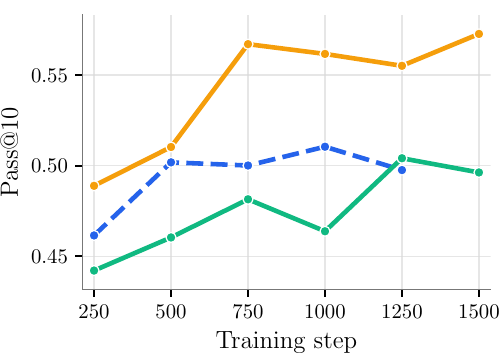}
        \caption{\passat{10}}
        \label{fig:baseline-dynamics-qwen4-taco-pass10}
    \end{subfigure}
    \hfill
    \begin{subfigure}[h]{0.385\textwidth}
        \centering
        \includegraphics[width=\linewidth]{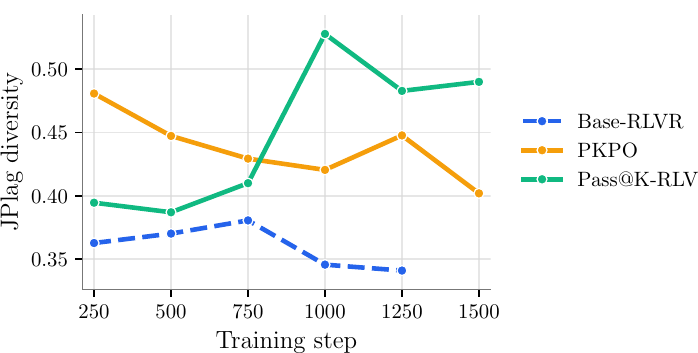}
        \caption{JPlag diversity}
        \label{fig:baseline-dynamics-qwen4-taco-diversity}
    \end{subfigure}
    \caption{Baseline training dynamics for Qwen3-4B on TACO-Cobalt.}
    \label{fig:baseline-dynamics-qwen4-taco}
\end{figure*}
\begin{figure*}[h]
    \centering
    \begin{subfigure}[h]{0.28\textwidth}
        \centering
        \includegraphics[width=\linewidth]{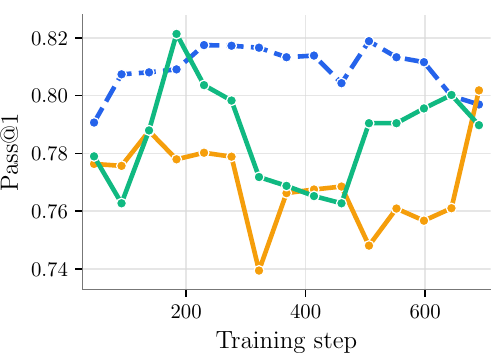}
        \caption{\passat{1}}
        \label{fig:baseline-dynamics-qwen8-mbpp-pass1}
    \end{subfigure}
    \hfill
    \begin{subfigure}[h]{0.28\textwidth}
        \centering
        \includegraphics[width=\linewidth]{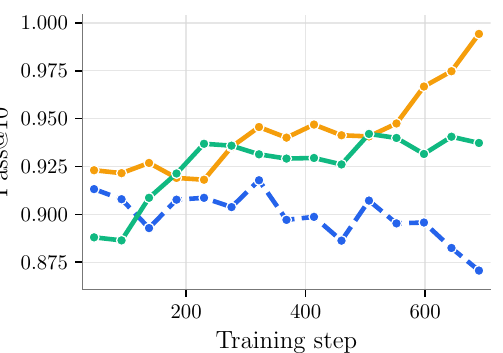}
        \caption{\passat{10}}
        \label{fig:baseline-dynamics-qwen8-mbpp-pass10}
    \end{subfigure}
    \hfill
    \begin{subfigure}[h]{0.385\textwidth}
        \centering
        \includegraphics[width=\linewidth]{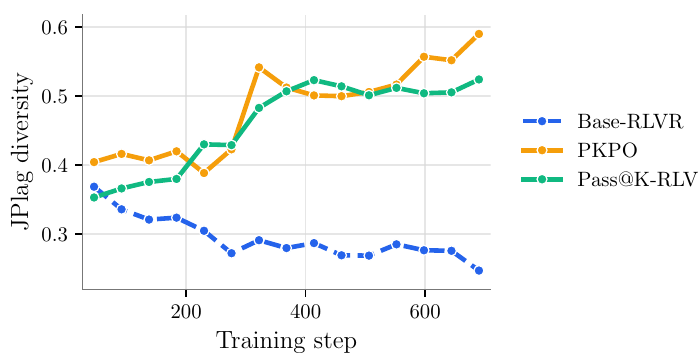}
        \caption{JPlag diversity}
        \label{fig:baseline-dynamics-qwen8-mbpp-diversity}
    \end{subfigure}
    \caption{Baseline training dynamics for Qwen3-8B on MBPP.}
    \label{fig:baseline-dynamics-qwen8-mbpp}
\end{figure*}
\begin{figure*}[h]
    \centering
    \begin{subfigure}[h]{0.28\textwidth}
        \centering
        \includegraphics[width=\linewidth]{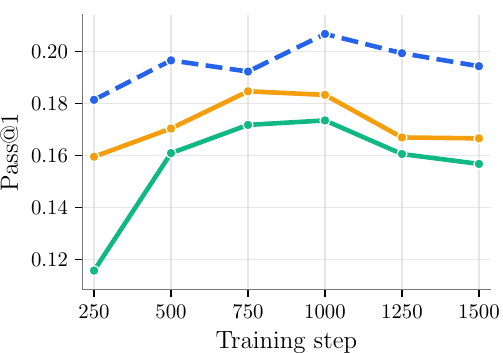}
        \caption{\passat{1}}
        \label{fig:baseline-dynamics-qwen8-code-contest-pass1}
    \end{subfigure}
    \hfill
    \begin{subfigure}[h]{0.28\textwidth}
        \centering
        \includegraphics[width=\linewidth]{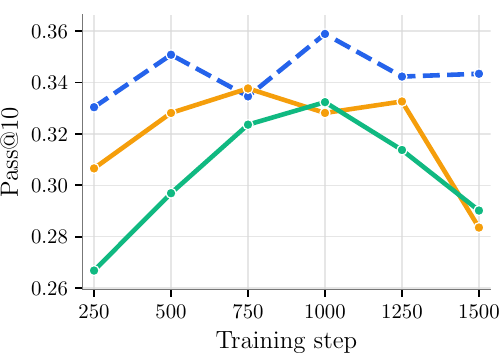}
        \caption{\passat{10}}
        \label{fig:baseline-dynamics-qwen8-code-contest-pass10}
    \end{subfigure}
    \hfill
    \begin{subfigure}[h]{0.385\textwidth}
        \centering
        \includegraphics[width=\linewidth]{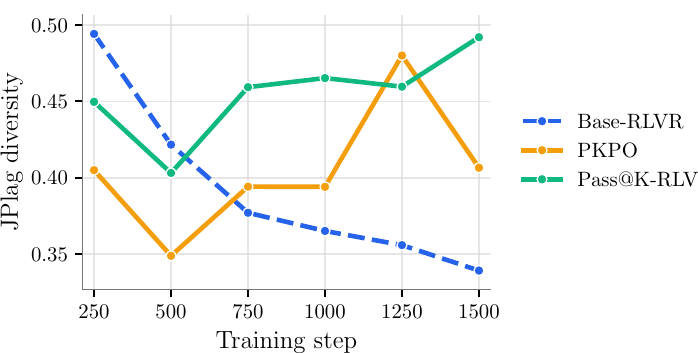}
        \caption{JPlag diversity}
        \label{fig:baseline-dynamics-qwen8-code-contest-diversity}
    \end{subfigure}
    \caption{Baseline training dynamics for Qwen3-8B on Code-Contest.}
    \label{fig:baseline-dynamics-qwen8-code-contest}
\end{figure*}
\begin{figure*}[h]
    \centering
    \begin{subfigure}[h]{0.28\textwidth}
        \centering
        \includegraphics[width=\linewidth]{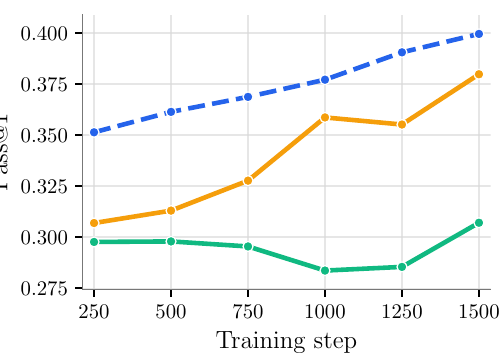}
        \caption{\passat{1}}
        \label{fig:baseline-dynamics-qwen8-taco-pass1}
    \end{subfigure}
    \hfill
    \begin{subfigure}[h]{0.28\textwidth}
        \centering
        \includegraphics[width=\linewidth]{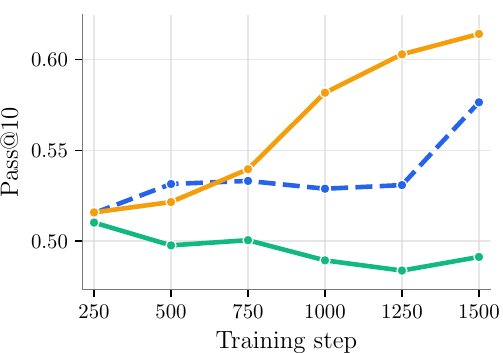}
        \caption{\passat{10}}
        \label{fig:baseline-dynamics-qwen8-taco-pass10}
    \end{subfigure}
    \hfill
    \begin{subfigure}[h]{0.385\textwidth}
        \centering
        \includegraphics[width=\linewidth]{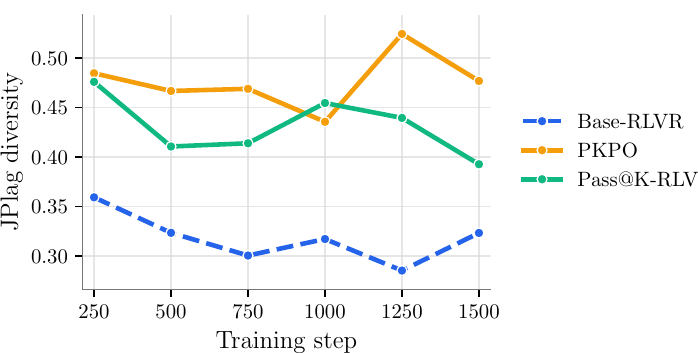}
        \caption{JPlag diversity}
        \label{fig:baseline-dynamics-qwen8-taco-diversity}
    \end{subfigure}
    \caption{Baseline training dynamics for Qwen3-8B on TACO-Cobalt.}
    \label{fig:baseline-dynamics-qwen8-taco}
\end{figure*}

Figures~\ref{fig:baseline-dynamics-olmo-code-contest}--\ref{fig:baseline-dynamics-qwen8-taco} provide the same three-panel view as Figure~\ref{fig:baseline-training-dynamics} for all model and dataset combinations. Together, these runs support the heterogeneous sample-redundancy pattern discussed in Section~\ref{sec:redundancy-analysis}: verifier training reliably improves executable correctness, while changing sampled-program redundancy.

The appendix trajectories cover Olmo3-7B on Code-Contest and TACO-Cobalt, plus Qwen3-4B and Qwen3-8B on MBPP, Code-Contest, and TACO-Cobalt. They show additional cases where executable performance and JPlag diversity evolve differently across baseline objectives.

